\documentclass{article}

\usepackage[final]{corl_2022}

\usepackage[utf8]{inputenc} %
\usepackage[T1]{fontenc}    %
\usepackage{url}            %
\usepackage{booktabs}       %
\usepackage{amsfonts}       %
\usepackage{nicefrac}       %
\usepackage{microtype}      %
\usepackage{tabularx}
\usepackage{microtype}
\usepackage{graphicx}
\usepackage{subfigure}
\usepackage{hyperref}
\hypersetup{
    colorlinks=true,
    linkcolor=blue,
    filecolor=magenta,      
    urlcolor=cyan,
}
\usepackage[utf8]{inputenc} %
\usepackage[T1]{fontenc}    %
\usepackage{dsfont}
\usepackage{url}            %
\usepackage{booktabs}       %
\usepackage{amsfonts}       %
\usepackage{nicefrac}       %
\usepackage{color}
\usepackage{xcolor}
\usepackage{mathtools}
\usepackage{amsmath,amssymb}
\usepackage{bm}
\usepackage{siunitx}
\usepackage{wrapfig}
\usepackage{algorithm}
\usepackage[noend]{algorithmic}
\usepackage{lipsum}
\usepackage{makecell}

\usepackage{amsmath}
\usepackage{amssymb}
\usepackage{mathtools}
\usepackage{amsthm}

\usepackage{amsmath,amsfonts,bm}

\def\eqref#1{equation~\ref{#1}}

\def\1{\bm{1}}

\DeclareMathAlphabet{\mathsfit}{\encodingdefault}{\sfdefault}{m}{sl}
\SetMathAlphabet{\mathsfit}{bold}{\encodingdefault}{\sfdefault}{bx}{n}

\DeclarePairedDelimiterX{\norm}[1]{\lVert}{\rVert}{#1}

\newcommand{\ie}{i.e., }
\newcommand{\eg}{e.g., }
\newcommand{\Skip}[1]{}

\newcommand{\acr}[0]{STAR}

\theoremstyle{plain}

\theoremstyle{definition}

\theoremstyle{remark}

\author{%
  Karl Pertsch$^1$\thanks{Work done during an internship at Meta AI. Correspondence to \href{mailto:pertsch@usc.edu}{\texttt{pertsch@usc.edu}}.}, \;\;Ruta Desai$^2$, \;\;Vikash Kumar$^2$, \\%
  \textbf{Franziska Meier$^2$, \;\;Joseph J. Lim$^3$, \;\;Dhruv Batra$^{2,4}$, \;\;Akshara Rai$^2$} \\[0.2cm]
  $^1$University of Southern California,\;$^2$Meta AI,\;$^3$KAIST,\;$^4$Georgia Tech\\
  \href{https://kpertsch.github.io/star}{\url{https://kpertsch.github.io/star}} 
}

\title{Cross-Domain Transfer via Semantic Skill Imitation}

\begin{document}

\maketitle
\begin{abstract}
We propose an approach for \emph{semantic} imitation, which uses demonstrations from a source domain, \eg human videos, to accelerate reinforcement learning (RL) in a different target domain, \eg a robotic manipulator in a simulated kitchen. Instead of imitating low-level actions like joint velocities, our approach imitates the sequence of demonstrated semantic skills like ``opening the microwave'' or ``turning on the stove''. This allows us to transfer demonstrations across environments (\eg real-world to simulated kitchen) and agent embodiments (\eg bimanual human demonstration to robotic arm). %
We evaluate on three challenging cross-domain learning problems and match the performance of demonstration-accelerated RL approaches that require in-domain demonstrations. %
In a simulated kitchen environment, our approach learns long-horizon robot manipulation tasks, using less than 3 minutes of human video demonstrations from a real-world kitchen. This enables scaling robot learning via the reuse of demonstrations, \eg collected as human videos, for learning in any number of target domains. %

\end{abstract}

\keywords{Reinforcement Learning, Imitation, Transfer Learning}

\section{Introduction}
\label{sec:intro}

\begin{wrapfigure}{r}{0.5\textwidth}
    \centering
    \vspace{-1.4cm}
    \includegraphics[width=\linewidth]{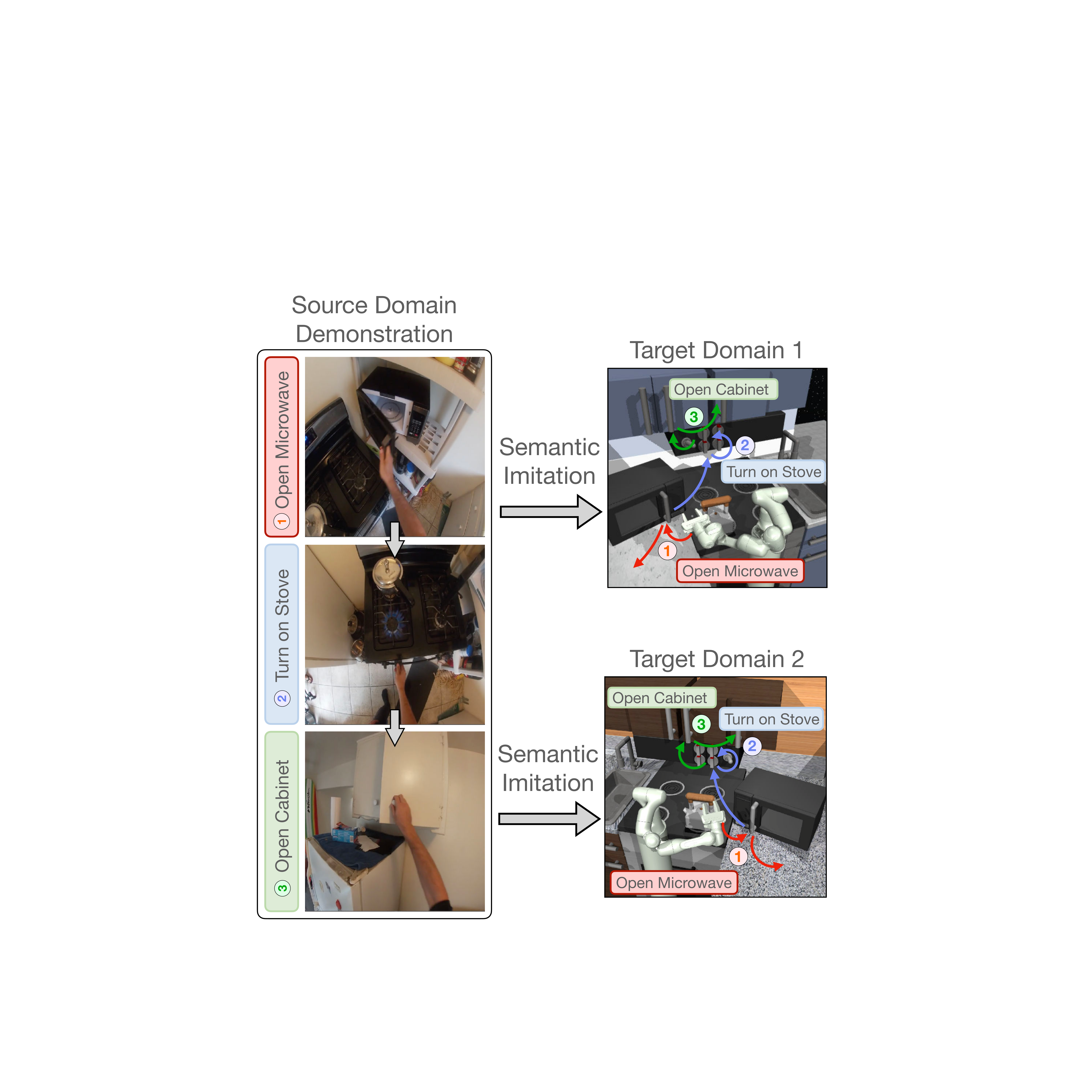}
    \caption{We address \emph{semantic} imitation, which aims to leverage demonstrations from a source domain, \eg human video demonstrations, to accelerate the learning of the same tasks in a different target domain, \eg controlling a robotic manipulator in a simulated kitchen environment. %
    }
    \label{fig:teaser}
    \vspace{-0.6cm}
\end{wrapfigure}

Consider a person imitating an expert in two scenarios: a beginner learning to play tennis, and a chef following a recipe for a new dish. In the former case, when mastering the basic skills of tennis, humans tend to imitate the precise arm movements demonstrated by the expert. %
In contrast, when operating in a familiar domain, such as a chef learning to cook a new dish, imitation happens on a higher scale. Instead of imitating individual movements, they follow high-level, semantically meaningful skills like ``stir the mixture'' or ``turn on the oven''. Such \emph{semantic skills} generalize across environment layouts, and allow humans to follow demonstrations across substantially different environments. %

Most works that leverage demonstrations in robotics imitate low-level actions. Demonstrations are typically provided by manually moving the robot \citep{sharma2018multiple} or via teleoperation \citep{mandlekar2018roboturk}. A critical challenge of this approach is scaling: demonstrations need to be collected in every new environment. On the other hand, imitation of high-level (semantic) skills has the promise of generalization: demonstrations can be collected in one kitchen and applied to any number of kitchens, eliminating the need to re-demonstrate in every new environment. Learning via imitation of high-level skills can lead to scalable and generalizable robot learning. %

In this work, we present \textbf{S}emantic \textbf{T}ransfer \textbf{A}ccelerated \textbf{R}L (\acr), which accelerates RL using cross-domain demonstrations by leveraging semantic skills, instead of low-level actions. %
We consider a setting with significantly different source and target environments. Figure~\ref{fig:teaser} shows an example: a robot arm learns to do a kitchen manipulation task by following a visual human demonstration from a different (real-world) kitchen. %
An approach that follows the precise arm movements of the human will fail due to embodiment and environment differences. Yet, by following the demonstrated semantic skills like ``open the microwave" and ``turn on the stove", our approach can leverage demonstrations \emph{despite} the domain differences. %
Like the chef in the above example, 
we use prior experience for enabling this semantic transfer. We assume access to datasets of prior experience collected across many tasks, in both the source and target domains. From this data, we learn semantic skills like ``open the microwave'' or ``turn on the stove''. Next, we collect demonstrations of the task in the source domain and find ``semantically similar'' states in the target domain. %
Using this mapping, we learn a policy to follow the demonstrated semantic skills in semantically similar states in the target domain. %

We present results on two semantic imitation problems in simulation and on real-to-sim transfer from human videos. %
In simulation, we test \acr~in: (1)~a maze navigation task across mazes of different layouts and (2)~a sequence of kitchen tasks between two variations of the FrankaKitchen environment~\citep{gupta2019relay}. In both tasks our approach matches the learning efficiency of methods with in-domain demonstrations, despite only using cross-domain demonstrations. %
Additionally, we show that a human demonstration video recorded within 3~minutes in a real-world kitchen can accelerate the learning of long-horizon manipulation tasks in the FrankaKitchen by hundreds of thousands of robot environment interactions. %

In summary, our contributions are twofold: (1)~we introduce \acr, an approach for cross-domain transfer via learned semantic skills, (2)~we show that \acr~can leverage demonstrations across substantially differing domains to accelerate the learning of long-horizon tasks.  %

\section{Related Work}
\label{sec:related_work}

\noindent{\textbf{Learning from demonstrations.}} Learning from Demonstrations (LfD, \citet{argall2009survey}) is a popular method for learning robot behaviors using demonstrations of the target task, often collected by human operators. Common approaches include behavioral cloning (BC, \citet{pomerleau1989alvinn}) %
and adversarial imitation approaches \citep{ho2016generative}. %
A number of works have proposed approaches for combining these imitation objectives with reinforcement learning \citep{rajeswaran2018learning,nair2018overcoming,zhu2018reinforcement,peng2018deepmimic}. However, all of these approaches require demonstrations in the target domain, %
limiting their applicability to new domains. %
In contrast, our approach imitates the demonstrations' \emph{semantic} skills and thus enables transfer \emph{across} domains.

\noindent{\textbf{Skill-based Imitation.}} Using temporal abstraction via skills has a long tradition in hierarchical RL~\citep{sutton1999options,bacon2017option,nachum2018data}. Skills have also been used for the \emph{imitation} of long-horizon tasks. \citet{pertsch2021skild,hakhamaneshi2021fist} learn skills from task-agnostic offline experience \citep{pertsch2020spirl,ajay2020opal} and imitate demonstrated skills instead of primitive actions. But, since the learned skills do not capture semantic information, they require demonstrations in the target domain. \citet{xu2018neural,huang2019neural} divide long-horizon tasks into subroutines, but struggle if the two domains requires a different sequence of subroutines, \eg if skill pre-conditions are not met in the target environment. %
Our approach is robust to such mismatches without requiring demonstrations in the target domain.

\noindent{\textbf{Cross-Domain Imitation.}} %
\citet{peng2020learning} assume a pre-specified mapping between source and target domain. %
\citep{smith2019avid,das2020model} leverage offline experience to learn mappings while~\citep{duan2017one,sharma2019third,yu2018one} rely on paired demonstrations. A popular goal is to leverage human videos for robot learning since they are easy to collect at scale. \citep{sermanet2018time,chen2021learning} learn reward functions from human demonstrations and \citet{schmeckpeper2020reinforcement} add human experience to an RL agent's replay buffer, but they only consider short-horizon tasks and rely on environments being similar. \citet{yu2018hierarchical} meta-learn cross-domain subroutines, but cannot handle different subroutines between source and target. Our approach imitates long-horizon tasks across domains, without a pre-defined mapping and is robust to different semantic subroutines.

\section{Problem Formulation}
\label{sec:problem_formulation}

We define a source environment $S$ and a target environment $T$. In the source environment, we have $N$ demonstrations $\tau^S_{1:N}$ with $\tau^S_i = \{s^S_0, a^S_0, s^S_1, a^S_1, \dots\}$ sequences of states $s^S$ and actions $a^S$. Our goal is to leverage these demonstrations to accelerate training of a policy $\pi(s^T)$ in the target environment, acting on target states $s^T$ and predicting actions $a^T$. $\pi(s^T)$ maximizes the discounted target task reward $J^T = \mathbb{E}_\pi \big[ \sum_{l=0}^{L-1} \gamma^l R(s^T_l, a^T_l) \big]$ for an episode of length $L$. 
We account for different state-action spaces $(s^S, a^S)$ vs. $(s^T, a^T)$ between source and target, but drop the superscript in the following sections, assuming that the context makes it clear whether we are addressing source or target states. In Section~\ref{sec:matching} we describe how we bridge this environment gap. %
Without loss of generality we assume that the source and target environments are substantially different; sequences of low-level actions %
that solve a task in the source environment \emph{do not} lead to high reward in the target environment. In the following we will also use the term \emph{domain} to refer to two environments with this property. %
Yet, we assume that the demonstrations show a set of \emph{semantic skills}, which when followed in the target environment can lead to task success. Here the term \emph{semantic skill} refers to a high-level notion of skill, like ``open the microwave'' or ``turn on the oven'', which is independent of the environment-specific low-level actions required to perform it. We further assume that both source and target environment allow for the execution of the same set of semantic skills.

Semantic imitation requires an agent to understand the semantic skills performed in the demonstrations. We use task-agnostic datasets $\mathcal{D}_S$ and $\mathcal{D}_T$ in the source and target domains to extract such semantic skills. Each $\mathcal{D}_i$ consists of state-action trajectories %
collected across a diverse range of prior tasks, \eg from previously trained policies or teleoperation, as is commonly assumed in prior work~\citep{pertsch2020spirl,ajay2020opal,pertsch2021skild,hakhamaneshi2021fist}. We also assume discrete semantic skill annotations $k_t \in \mathcal{K}$, denoting the skill being executed at time step $t$. %
These can be collected manually, but we demonstrate how to use pre-trained action recognition models as a more scalable alternative (Sec.~\ref{sec:human-demo}).

\section{Approach}
\label{sec:approach}
\begin{wrapfigure}{r}{0.6\textwidth}
\vspace{-2.0cm}
\begin{minipage}{0.6\textwidth}
\begin{algorithm}[H]
\caption{\acr~(Semantic Transfer Accelerated RL)}
\label{alg:sil_summary}
\begin{algorithmic}

\STATE Pre-Train low-level policy $\pi^l(a \vert s, k, z)$ \COMMENT{cf. Sec.~\ref{sec:skill_model}}
\STATE Match source demos to target states \COMMENT{cf. Sec.~\ref{sec:matching}}
\STATE Pre-train $p^\text{demo}(k \vert s), p^\text{TA}(k \vert s), p^\text{TA}(z \vert s, k), D(s)$ \COMMENT{cf. Tab.~\ref{tab:models_and_objectives}}

\FOR{each target train iteration} 
\STATE Collect online experience $(s, k, z, R, s^\prime$)
\STATE Update high-level policies with eq.~\ref{eq:policy_objective} \COMMENT{cf. Alg.~\ref{alg:sil}}

\ENDFOR
\STATE \textbf{return} trained high-level policies $\pi^\text{sem}(k \vert s), \pi^\text{lat}(z \vert s, k)$
\end{algorithmic}
\end{algorithm}
\end{minipage}
\vspace{-0.4cm}
\end{wrapfigure}

Our approach \acr~imitates demonstrations' semantic skills, instead of low-level actions, to enable cross-domain, \emph{semantic} imitation. We use a two-layer hierarchical policy with a high-level that outputs the semantic skill and a low-level that executes the skill. We first describe our semantic skill representation, followed by the low-level and high-level policy learning.
Algorithm~\ref{alg:sil_summary} summarizes our approach. %

\subsection{Semantic Skill Representation}
\label{sec:skill_model}

\begin{figure}[t]
  \centering
  \includegraphics[width=\linewidth]{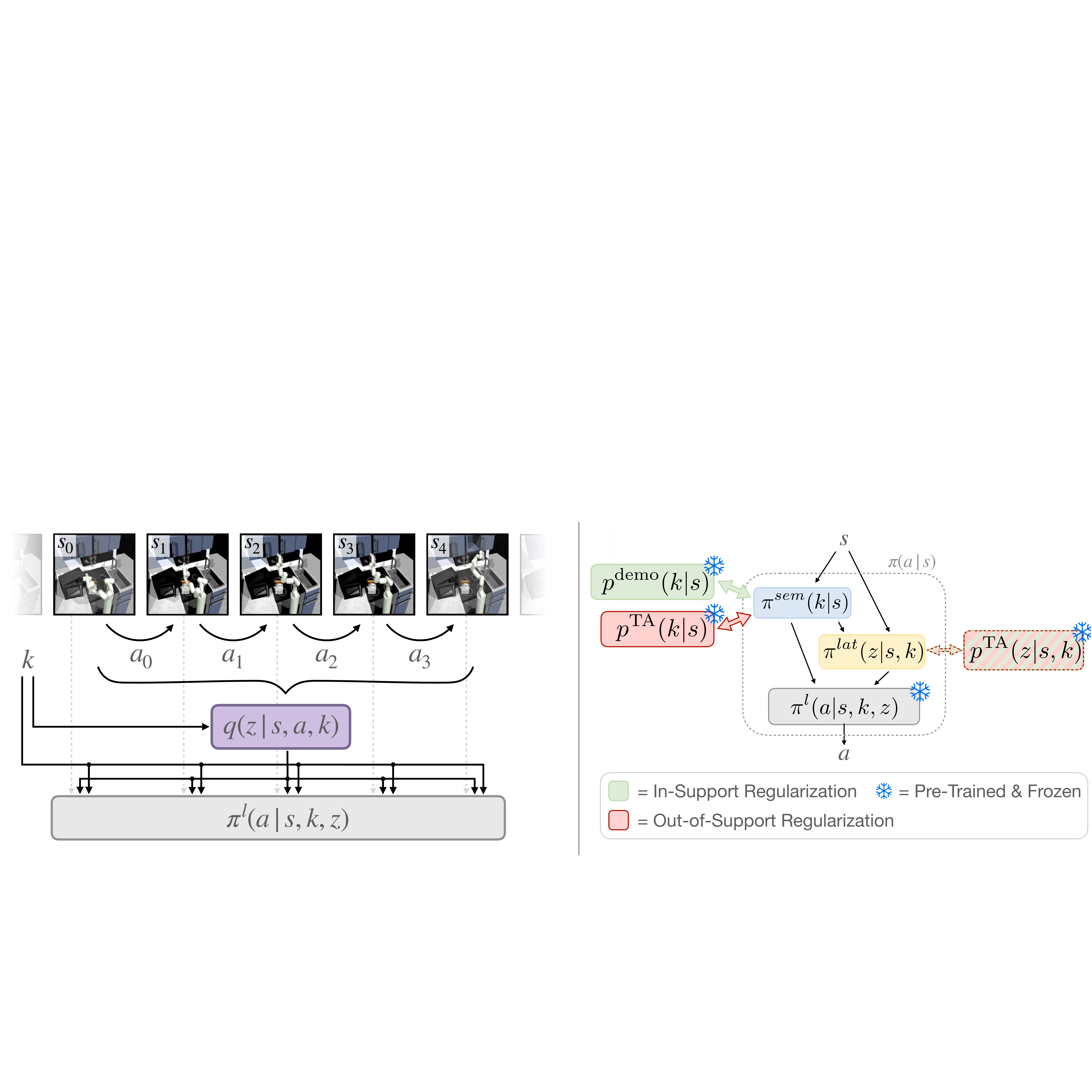}
  \vspace{-0.5cm}
  \caption{
  Model overview for pre-training (\textbf{left}) and target task learning (\textbf{right}). We pre-train a semantic skill policy $\pi^l$ (\textcolor[HTML]{929292}{\textbf{grey}}) and use it to decode actions from the learned high-level policies $\pi^\text{sem}$ and $\pi^\text{lat}$ (\textcolor[HTML]{0F52BA}{\textbf{blue}} and \textcolor[HTML]{F4C430}{\textbf{yellow}}) during target task learning. See training details in the main text.}
    \label{fig:model_and_policy}
\end{figure}

A skill is characterized by both its semantics, \ie whether to open the microwave or turn on the stove, as well as the details of its low-level execution, \eg at what angle to approach the microwave or where to grasp its door handle. %
Thus, we represent skills via a low-level policy $\pi^l(a \vert s, k, z)$ %
which is conditioned on the current environment state $s$, the semantic skill ID $k$ and a latent variable $z$ which captures the execution details. 
For example, when ``turning on the stove", $a$ are the joint velocities, $s$ is the robot and environment state, $k$ is the semantic skill ID of this skill, and $z$ captures the robot hand orientation as it interacts with the stove. %
A single tuple $(k, z)$ represents a sequence of $H$ steps, since such temporal abstraction facilitates long-horizon imitation~\citep{pertsch2021skild}. We train our model as a conditional variational autoencoder (VAE)~\citep{sohn2015learning} over a sequence of actions \emph{given} a state and semantic skill ID. Thus, the latent variable $z$ represents all information required to reconstruct $a_{0:H-1}$ that is not contained in the skill ID, \ie information about \emph{how} to execute the semantic skill.

Figure~\ref{fig:model_and_policy}, left depicts the training setup for $\pi^l$. We randomly sample an $H$-step state-action subsequence $(s_{0:H}, a_{0:H-1})$ from $\mathcal{D}_T$. An inference network $q(z \vert s, a, k)$ encodes the sequence into a latent representation $z$ conditioned on the semantic skill ID $k$ at the first time step. %
$k$ and $z$ are passed to $\pi^l$, which reconstructs the sampled actions. %
Our training objective is a standard conditional VAE objective that combines a reconstruction and a prior regularization term:
\begin{equation}
    \mathcal{L}_{\pi_l} = \underbrace{\mathbb{E}_q \bigg[\prod_{t=0}^{H-1} \log \pi^l(a_t \vert s_t, k, z) \bigg]}_{\text{reconstruction}} - \underbrace{\beta D_\text{KL} \big(q(z \vert s_{0:H}, a_{0:H-1}, k), p(z)\big)}_{\text{prior regularization}} .
\label{eq:model_objective}
\end{equation}

Here $D_\text{KL}$ denotes the Kullback-Leibler divergence. We use a simple uniform Gaussian prior $p(z)$ and a weighting factor $\beta$ for the regularization objective~\citep{higgins2017beta}. %
The semantic skill ID $k$ is pre-defined, discrete and labelled, while the latent $z$ is learned and continuous. In this way, our formulation captures discrete aspects of manipulation skills (open a microwave vs. turn on a stove) while being able to continuously modulate each semantic skill (\eg different ways of approaching the microwave). %

\subsection{Semantic Transfer Accelerated RL}
\label{sec:star_core}

After pre-training the low-level policy $\pi^l(a \vert s, k, z)$, we learn the high-level policy using the source domain demonstrations. %
Concretely, we train a policy $\pi^{h}(k, z \vert s)$ that predicts tuples $(k, z)$ which get executed via $\pi^l$. Note that unlike prior work \citep{pertsch2021skild}, our high-level policy outputs both, the semantic skill $k$ \emph{and} the low-level execution latent $z$. It is thus able to choose which semantic skill to execute and tailor its execution to the target domain. Cross-domain demonstrations solely guide the \emph{semantic} skill choice, since the low-level execution might vary between source and target domains. Thus, we factorize $\pi^h$ into a semantic sub-policy $\pi^\text{sem}(k \vert s)$ and a latent, non-semantic sub-policy $\pi^\text{lat}(z \vert s, k)$:
\begin{equation}
    \pi(a \vert s) = \underbrace{\pi^l(a \vert s, k, z)}_\text{skill policy} \cdot \underbrace{\pi^\text{lat}(z \vert s, k) \; \pi^\text{sem}(k \vert s)}_\text{high-level policy $\pi^{h}(k, z \vert s)$}. 
\end{equation}
Intuitively, this can be thought of as first deciding \textit{what} skill to execute (\eg open the microwave), followed by \textit{how} to execute it. %
We pre-train multiple models via supervised learning for training $\pi^{h}$: (1)~two semantic skill priors $p^{\text{demo}}(k \vert s)$ and $p^{\text{TA}}(k \vert s)$, trained to infer the semantic skill annotations from demonstrations and task-agnostic dataset $\mathcal{D}_T$ respectively, (2)~a task-agnostic prior $p^{\text{TA}}(z \vert s, k)$ over the latent skill variable $z$, trained to match the output of the inference network on $\mathcal{D}_T$ and (3)~a discriminator $D(s)$, trained to classify whether a state is part of the demonstration trajectories. We summarize all pre-trained components and their \emph{supervised} training objectives in Appendix, Table~\ref{tab:models_and_objectives}.

We provide an overview of our semantic imitation architecture and the used regularization terms in Figure~\ref{fig:model_and_policy}, right. We build on the idea of weighted policy regularization with a learned demonstration support estimator from~\citet{pertsch2021skild} (for a brief summary, see appendix~\ref{sec:skild_summary}). We regularize the high-level \emph{semantic} policy $\pi^\text{sem}$ (\textcolor[HTML]{0F52BA}{\textbf{blue}}) towards the demonstration skill distribution $p^{\text{demo}}(k \vert s)$ when $D(s)$ classifies the current state as part of the demonstrations (\textcolor[HTML]{00A86B}{\textbf{green}}). %
For states which $D(s)$ classifies as outside the demonstration support, we regularize $\pi^\text{sem}$ towards the task-agnostic prior $p^{\text{TA}}(k \vert s)$ (\textcolor[HTML]{FF9090}{\textbf{red}}). We \emph{always} regularize the non-semantic sub-policy $\pi^\text{lat}(z \vert s, k)$ (\textcolor[HTML]{F4C430}{\textbf{yellow}}) towards the task-agnostic prior $p^\text{TA}(z \vert s, k)$, since execution-specific information cannot be transferred across domains.
The overall optimization objective for $\pi^{h}$ is:
\begin{align}
    \mathbb{E}_{\pi^h} \bigg[\tilde{r}(s, a) 
    &\underbrace{- \alpha_q D_\text{KL}\big(\pi^\text{sem}(k \vert s), p^{\text{demo}}(k \vert s)\big) \cdot D(s)}_{\text{demonstration regularization}}
    \underbrace{- \alpha_p D_\text{KL}\big(\pi^\text{sem}(k \vert s), p^{\text{TA}}(k \vert s)\big) \cdot (1-D(s))}_{\text{task-agnostic semantic prior regularization}}, \nonumber\\
    &\underbrace{- \alpha_l D_\text{KL}\big(\pi^\text{lat}(z \vert s, k), p^\text{TA}(z \vert s, k)\big)}_{\text{task-agnostic execution prior regularization}}\bigg].
\label{eq:policy_objective}
\end{align}
$\alpha_q$, $\alpha_p$ and $\alpha_l$ are either fixed or automatically tuned via dual gradient descent. We augment the target task reward using the discriminator $D(s)$ to encourage the policy to reach states within the demonstration support: $\tilde{r}(s, a) = (1-\kappa) \cdot R(s, a) + \kappa \cdot \big[\log D(s) - \log\big(1 - D(s)\big)\big]$. In the setting with no target environment rewards (pure imitation learning), we rely solely on this discriminator reward for policy training (Section~\ref{sec:imitation_results}). For a summary of the full procedure, see Algorithm~\ref{alg:sil}.

\Skip{
The task-agnostic prior $p^{\text{TA}}(k \vert s)$ is especially important in \emph{cross-domain} transfer, since the \textit{exact} sequence of semantic skills from the source domain might not be feasible, and the high-level policy might have to take detours. For example, consider a scenario in which the demonstrator already has a pot on the kitchen counter and pours water into it. Imitating the same in your kitchen might require you to first retrieve a pot. Such scenarios make rigidly following the semantic skill sequence from the demonstration prone to failure, unless source and target environments are arranged to be very close. On the other hand, our approach is robust to such mismatches between source and target by determining if a state is within demonstration support, and only following the demonstrated skills when it is appropriate. In mismatched settings, the task-agnostic prior helps the policy explore the environment to discover missing steps in the demonstration, until the state is within demonstration support again.  
}

The final challenge is that the discriminator $D(s)$ and the prior $p^{\text{demo}}(k \vert s)$ are trained on states from the source domain, but need to be applied to the target domain. %
Since the domains differ substantially, we cannot expect the pre-trained networks to generalize. Instead, we need to explicitly bridge the state domain gap, as described next. %

\subsection{Cross-Domain State Matching}
\label{sec:matching}

\begin{wrapfigure}{r}{0.5\textwidth}
    \centering
    \vspace{-1.7cm}
    \includegraphics[width=0.9\linewidth]{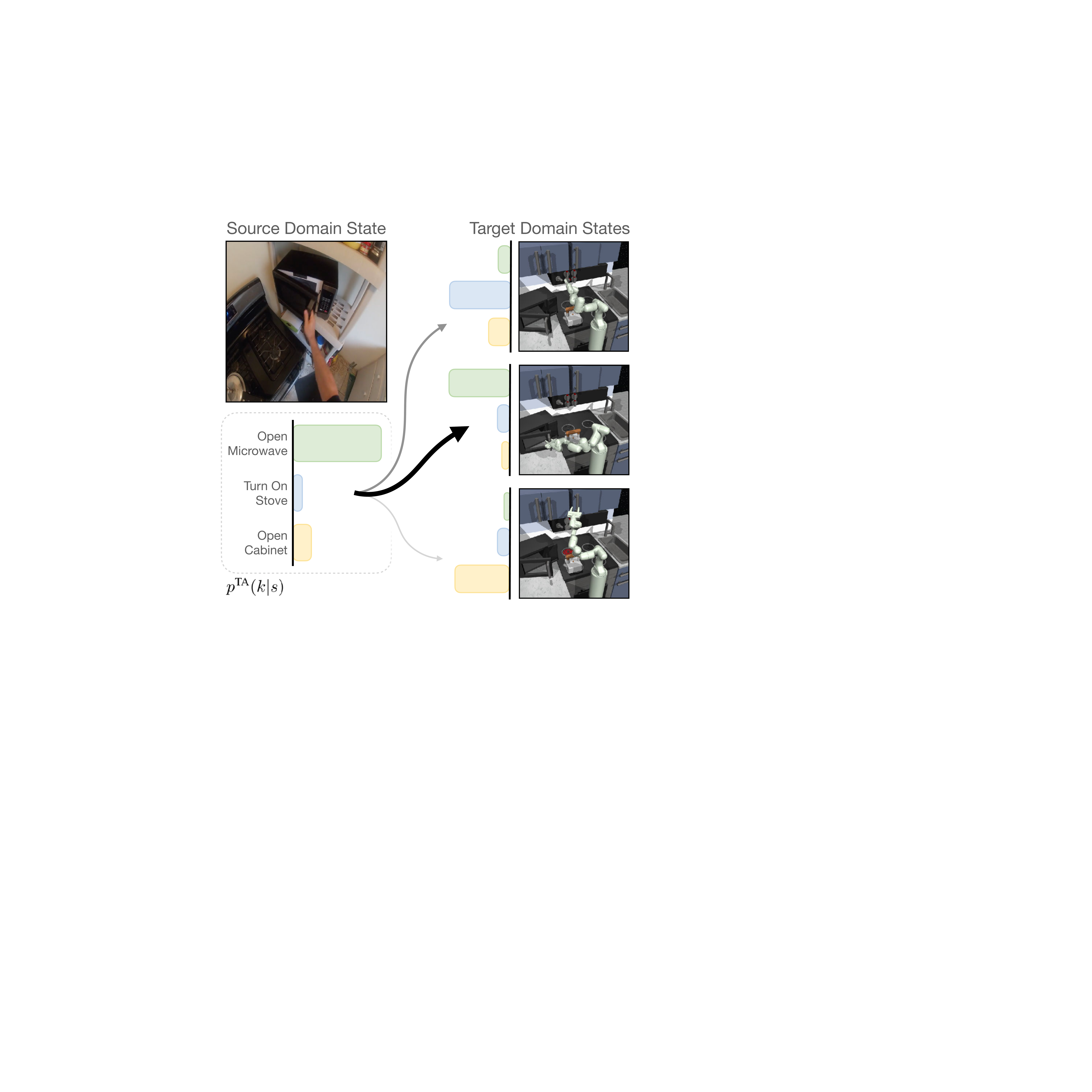}
    \caption{State matching between source and target domain. For every source domain state from the demonstrations, we compute the task-agnostic semantic skill distribution $p^\text{TA}(k \vert s)$ and find the target domain state with the most similar semantic skill distribution from the task-agnostic dataset $\mathcal{D}_T$. We then relabel the demonstrations with these matched states from the target domain. 
    }
    \label{fig:matching}
    \vspace{-0.6cm}
\end{wrapfigure}

Our goal is to find semantically similar states between the source and the target environment. These are states with similar distributions over likely semantic skills. E.g. if the agent’s hand is reaching for the handle of a closed microwave, the probability for the skill “open microwave” is high, while the probability for other skills, e.g. “turn on stove” is low. Crucially, this is true independent of the domain and independent of whether e.g. a human or robot is executing the action. Thus, we can use the skill prior distributions to find \emph{semantically} similar states.

Following this intuition, we find corresponding states based on the similarity between the task-agnostic semantic skill prior distributions $p^{\text{TA}}(k \vert s)$. We illustrate an example in Figure~\ref{fig:matching}: for a given source demonstration state $s^S$ with high likelihood of opening the microwave, we find a target domain state $s^T$ that has high likelihood of opening the microwave, by minimizing the symmetric KL divergence between the task-agnostic skill distributions %
(we omit $(\cdot) ^{\text{TA}}$ for brevity):
\begin{align}
    \vspace{-0.1cm}
\label{eq:matching}
    \min_{s^T \in \mathcal{D}_T} D_\text{KL}\big(&p_T(k \vert s^T), p_S(k \vert s^S)\big) 
    + D_\text{KL}\big(p_S(k \vert s^S), p_T(k \vert s^T)\big)
    \vspace{-0.3cm}
\end{align}

In practice, states can be matched incorrectly %
when the task agnostic dataset chooses one skill with much higher probability than others. In such states, the divergence in \eqref{eq:matching} is dominated by one skill, and others are ignored, causing matching errors. Using a state's temporal context can result in more robust correspondences by reducing the influence of high likelihood skills in any single state. We compute an aggregated skill distribution $\phi(k \vert s)$ using a temporal window around the current state:
\begin{equation}
    \phi(k \vert s_t) = \frac{1}{Z(s)} \bigg( \sum_{i=t}^T \gamma_+^i p(k \vert s_i) + \sum_{j=1}^{t-1} \gamma_{-}^{t-j} p(k \vert s_{t-j})\bigg) 
\end{equation}
Here, $\gamma_{+}, \gamma_{-} \in [0, 1]$ determine the forward and backward horizon of the aggregate skill distribution. $Z(s)$ ensures that the aggregate probability distribution sums to one. Instead of $p^\text{TA}$ in~\eqref{eq:matching}, we use $\phi(k \vert s)$.
By matching all source-domain demonstrations states to states in the target domain via $\phi(k \vert s)$, we create a proxy dataset of target state demonstrations, which we use to pre-train the models $p^{\text{demo}}(k \vert s)$ and $D(s)$. Once trained, we use them for training the high-level policy via \eqref{eq:policy_objective}.

\section{Experiments}
\label{sec:experiments}
\begin{wrapfigure}{r}{0.6\textwidth}
\vspace{-1.5cm}
    \centering
    \includegraphics[width=1.0\linewidth]{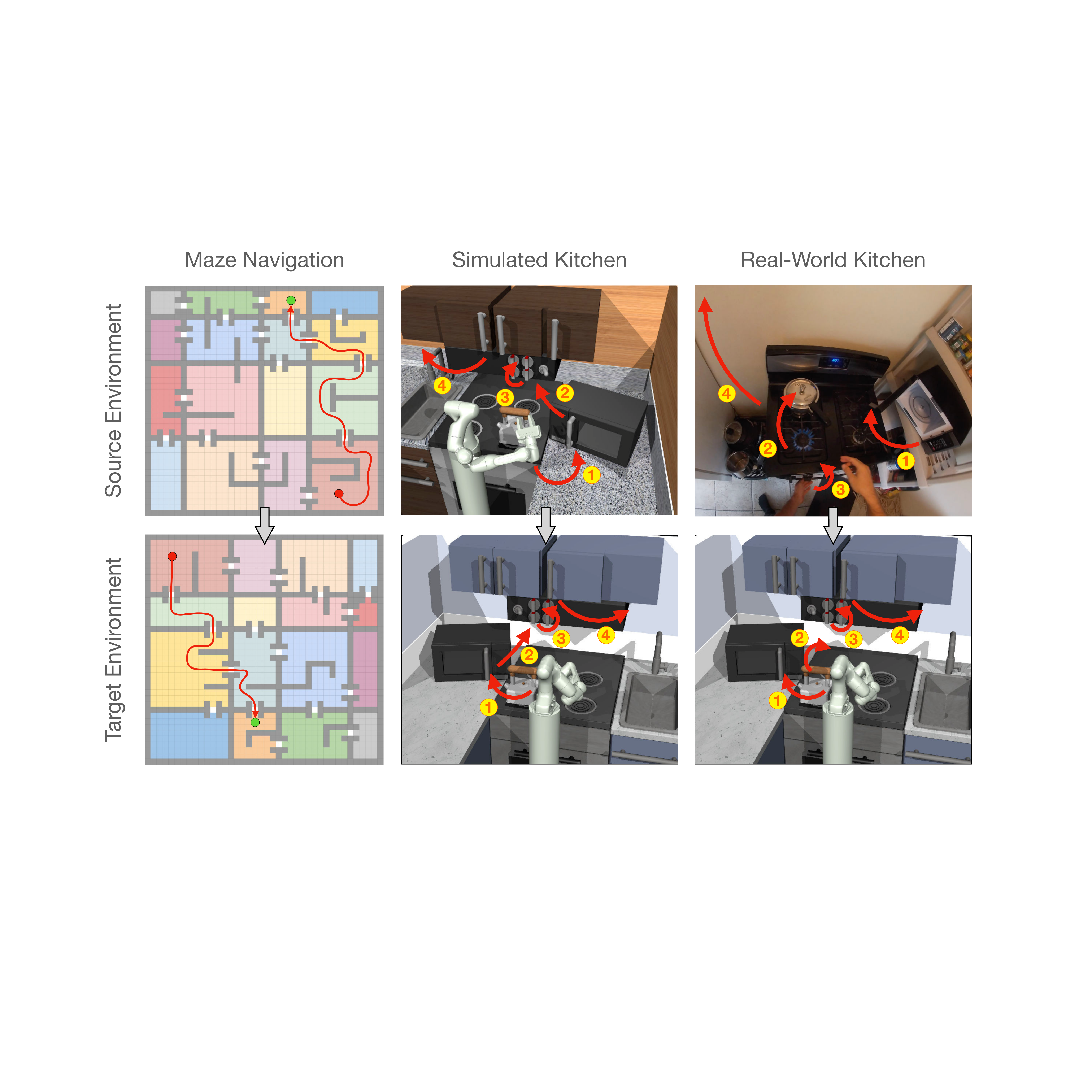}
    \caption{We evaluate on three pairs of source (\textbf{top}) and target (\textbf{bottom}) environments. \textbf{Left}: maze navigation. The agent needs to follow a sequence of colored rooms (red path) but the maze layout changes substantially between source and target domains. \textbf{Middle}: kitchen manipulation. A robotic arm executes a sequence of skills, but the layout of the kitchens differs. \textbf{Right}: Same as before, but with human demonstrations from a real-world kitchen.
    }
    \label{fig:environments}
        \vspace{-.8cm}
\end{wrapfigure}
Our experiments are designed to answer the following questions: (1)~Can we leverage demonstrations \emph{across domains} to accelerate learning via semantic imitation? %
(2)~Can we use semantic imitation to teach a robot a new task from real-world videos of humans performing the task? (3)~Is our approach %
robust to missing skills in the demonstrations?
We test semantic imitation across two simulated maze and kitchen environments, as well as from real-world videos of humans to a simulated robot. Our results show that our approach can accelerate learning from cross-domain demonstrations, even with real-to-sim gap.%

\subsection{Cross-Domain Imitation in Simulation} %
\label{sec:si_results}

We first test our approach \acr~in two simulated settings: %
a maze navigation and a robot kitchen manipulation task (see Figure~\ref{fig:environments}, left \& middle). In the \textbf{maze navigation task}, both domains have corresponding rooms, indicated by their color in Figure~\ref{fig:environments}. The agent needs to follow a sequence of semantic skills like ``go to red room'', ``go to green room'' etc. %
In the \textbf{kitchen manipulation task}, a Franka arm tackles long-horizon manipulation tasks in a simulated kitchen~\citep{gupta2019relay}. We define 7 semantic skills, like ``open the microwave'' or ``turn on the stove'' in the source and target environments. %
In both environments we collect demonstrations in the source domain, and task-agnostic datasets in both the source and target domains using motion planners and human teleoperation respectively. For further details on action and observation spaces, rewards and data collection, see Sec~\ref{sec:details_env_and_data}.

\Skip{
\noindent{\textbf{Data Collection:}} We collect task demonstrations in the source domains, and task-agnostic datasets in both the source and target domains. In the maze navigation tasks, we use the planner from \citet{fu2020d4rl} to generate 9000 task-agnostic sequences between randomly sampled start-goal pairs as well as 100 demonstrations in the source domain (marked in red in Figure~\ref{fig:environments}). In the simulated kitchen domain, we use the dataset of \citet{gupta2019relay} as our task-agnostic dataset, consisting of 600 teleoperated sequences in which the agent performs diverse semantic skills. In the source domain, we use a CEM-based planner~\citep{blossom2006cross} for generating 600 task-agnostic sequences and 20 demonstrations of the target task (see Figure~\ref{fig:environments}).
}

We compare our approach to multile prior skill-based RL approaches with and without demonstration guidance: \textbf{SPiRL}~\citep{pertsch2020spirl} learns skills from $\mathcal{D}_T$ and then trains a high-level policy over skills; \textbf{BC+RL}~\citep{rajeswaran2018learning,nair2018overcoming} pre-trains with behavioral cloning and finetunes with SAC~\citep{haarnoja2018sac}; \textbf{SkillSeq}, similar to \citet{xu2018neural}, sequentially executes the ground truth sequence of semantic skills as demonstrated; \textbf{SkiLD}~\citep{pertsch2021skild} is an oracle with access to demonstrations \emph{in the target domain} and follows them using learned skills. For more details on the implementation of our approach and all comparisons, see appendix, Sections~\ref{sec:impl_details_skill_learning}~-~\ref{sec:impl_details_comparisons}.

\Skip{
\noindent{\textbf{Comparisons:}} First, we compare \acr~to a baseline that only uses the task-agnostic data but no demonstrations: \textbf{SPiRL}~\citep{pertsch2020spirl} learns skills from $\mathcal{D}_T$ and then trains a high-level policy over skills. This comparison shows how cross-domain demonstrations can improve the learning efficiency over no demonstrations. Next, we compare to a standard Behaviour cloning and RL (\textbf{BC+RL}) baseline~\citep{rajeswaran2018learning,nair2018overcoming}, to demonstrate that the substantial domain difference makes direct imitation of the low-level actions unsuitable for learning. We upper bound the performance of our approach using an oracle baseline, \textbf{SkiLD}~\citep{pertsch2021skild}, which has access to demonstrations \emph{in the target domain} and follows them using learned skills. %
Finally, we compare to an alternative semantic imitation approach, \textbf{SkillSeq}, similar to \citet{xu2018neural}, which sequentially executes the semantic skills as demonstrated. Policies for each semantic skill as well as termination predictors are learned from the task-agnostic target domain data. For fair comparison, we fine-tune \textbf{SkillSeq} using SAC~\citep{haarnoja2018sac}. %
For more details on environment setup and the implementation of our approach and all comparisons, see appendix, Section~\ref{sec:impl_details}.
}

\begin{figure}[t]
    \centering
    \includegraphics[width=1.0\linewidth]{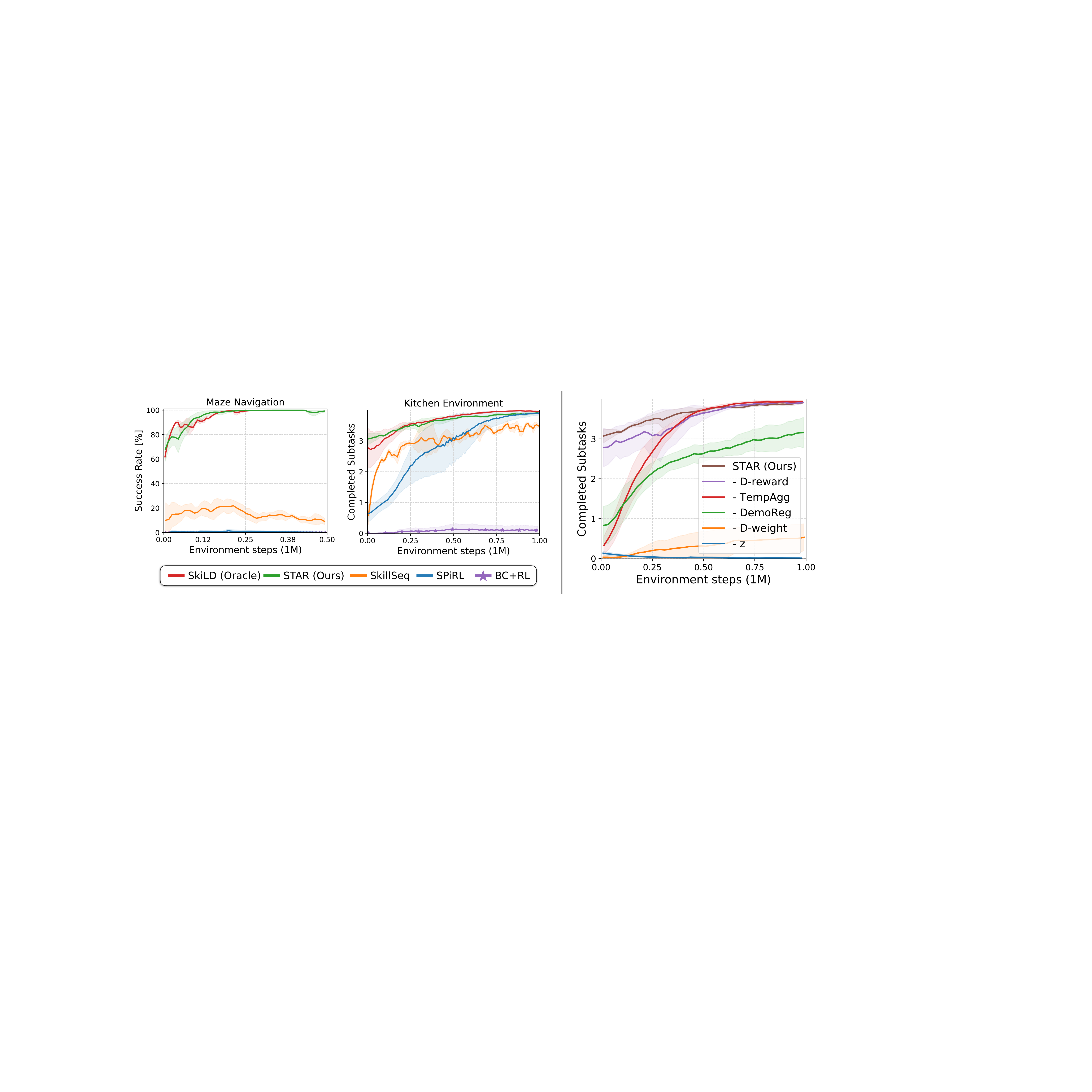}
    \vspace{-0.5cm}
    \caption{\textbf{Left}: Performance on the simulated semantic imitation tasks. \acr, matches the performance of the oracle, SkiLD, which has access to target domain demonstrations and outperforms both SPiRL, which does not use demonstrations, and SkillSeq, which follows the demonstrated semantic skills sequentially. %
    \textbf{Right}: Ablations in the kitchen environment, see main text for details.
    }
    \label{fig:quant_results}
\end{figure}

Figure~\ref{fig:quant_results}, left, compares the performance of all approaches in both tasks. BC+RL is unable to leverage the cross-domain demonstrations and makes no progress on the task. %
SPiRL %
is able to learn the kitchen manipulation task, but requires many more environment interactions to reach the same performance as our approach. %
SkillSeq succeeds in approximately 20\% of the maze episodes and solves on average 3 out of 4 subtasks in the kitchen manipulation environment after fine-tuning. The mixed success is due to inaccuracies in execution of the skill policies since SkillSeq follows the ground truth sequence of high-level skills. %
Our approach, \acr, can use cross-domain demonstrations to match the learning efficiency of SkiLD (oracle) that has access to \textit{target domain} demonstrations. This shows that our approach %
is effective at extracting useful information from cross-domain demonstrations. During downstream task training of the high-level semantic and execution policies our approach can fix both, errors in the high-level skill plan and the low-level skill execution. The ability to jointly adapt high-level and low-level policies and e.g. react to failures in the low-level policy rather than following a fixed high-level plan is crucial for good performance on long-horizon tasks. %
We find that this trend holds even in the ``pure'' imitation learning (IL) setting without environment rewards, where we solely rely on the learned discriminator reward to guide learning (see appendix, Section~\ref{sec:imitation_results} for detailed results). %
Thus, \acr~can be used both, as a demonstration-guided RL algorithm and for cross-domain imitation learning. 
Qualitative results can be viewed at {\url{https://tinyurl.com/star-rl}} and in Figure~\ref{fig:quali_maze_results}. 

To study the different components of our approach, we run ablations in the FrankaKitchen environment (Fig.~\ref{fig:quant_results}, right). Removing the discriminator-based weighting for the demonstration regularization (\textbf{-D-weight}) (Eq. 4) or removing the demonstration regularization altogether (\textbf{-DemoReg}), leads to poor performance. In contrast, removing the discriminator-based dense reward (\textbf{-D-reward}) or temporal aggregation during matching (\textbf{-TempAgg}) affects learning speed but has the same asymptotic performance. Finally, a model without the latent variable $z$ (\textbf{-z}) cannot model the diversity of skill executions in the data; the resulting skills are too imprecise to learn long-horizon tasks. 
We show qualitative examples of the effect of varying matching window sizes $[\gamma^-, \gamma^+]$ on the project website: {\url{https://tinyurl.com/star-rl}}.

\subsection{Imitation from Human Demonstrations}
\label{sec:human-demo}

\begin{figure}[t]
    \centering
    \includegraphics[width=\linewidth]{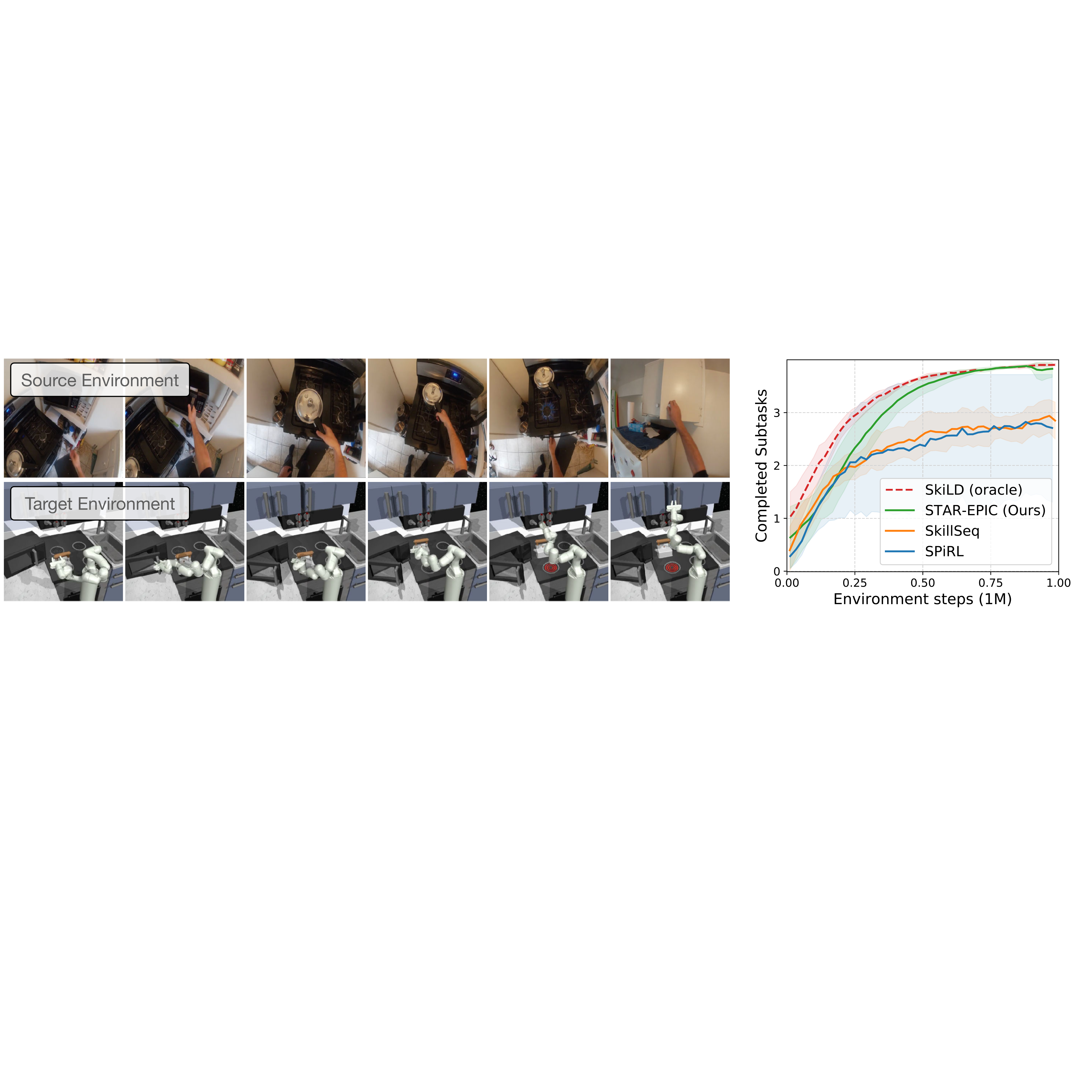}
    \caption{Semantic imitation from human demonstrations. \textbf{Left}: Qualitative state matching results. The top row displays frames subsampled from a task demonstration in the human kitchen source domain. The bottom row visualizes the states matched to the source frames via the procedure described in Section~\ref{sec:matching}. The matched states represent corresponding \emph{semantic} scenes in which the agent \eg opens the microwave, turns on the stove or opens the cabinet. 
    \textbf{Right}: Quantitative results on the kitchen manipulation task from human video demonstrations.
    }
    \label{fig:human_results}
    \vspace{-0.6cm}
\end{figure}

In this section we ask: can our approach be used to leverage human video demonstrations for teaching new tasks to robots? Imitating human demonstrations presents a larger challenge since it requires bridging domain differences that span observation spaces (from images in the real-world to low-dimensional states in simulation), agent morphologies (from a bimanual human to a 7DOF robot arm), and environments (from the real-world to a simulated robotic environment).
To investigate this question, we collect 20 human video demonstrations in a real-world kitchen, which demonstrate a task the robotic agent needs to learn in the target simulated domain. Instead of collecting a large, task-agnostic dataset in the human source domain and manually annotating semantic skill labels, we demonstrate a more scalable alternative: we use an action recognition model, pre-trained on the EPIC Kitchens dataset~\citep{Damen2021RESCALING}, zero-shot to predict semantic skill distributions on the human demonstration videos. We define a mapping from the 97 verb and 300 noun classes in EPIC Kitchens to the skills present in the target domain and then use our approach as described in Section~\ref{sec:star_core}, using the EPIC skill distributions as the task-agnostic skill prior $p^\text{TA}(k \vert s)$. For data collection details, see Section~\ref{sec:details_env_and_data}.

We visualize qualitative matching results between the domains in Figure~\ref{fig:human_results}, left. We successfully match frames to the corresponding \emph{semantic} states in the target domain. %
In Figure~\ref{fig:human_results}, right, we show that this leads to successful semantic imitation of the human demonstrations. Our approach \acr~with EPIC Kitchens auto-generated skill distributions is able to reach the same asymptotic performance as the oracle approach that has access to target domain demonstrations, with only slightly reduced learning speed. It also outperforms the SkillSeq and SPiRL baselines %
(for qualitative results see %
\url{https://tinyurl.com/star-rl}). %

To recap: for this experiment we \emph{did not} collect a large, task-agnostic human dataset and we \emph{did not} manually annotate any human videos. Collecting a few human demonstrations in an unseen kitchen was sufficient to substantially accelerate learning of the target task on the robot in simulation. This demonstrates one avenue for scaling robot learning by (1) learning from easy-to-collect human video demonstrations and (2) using pre-trained skill prediction models to bridge the domain gap.

\subsection{Robustness to Noisy Demonstrations and Labels}
\label{sec:robustness_exp}
\begin{wrapfigure}{r}{0.6\textwidth}
    \centering
    \vspace{-.5cm}
    \includegraphics[width=\linewidth]{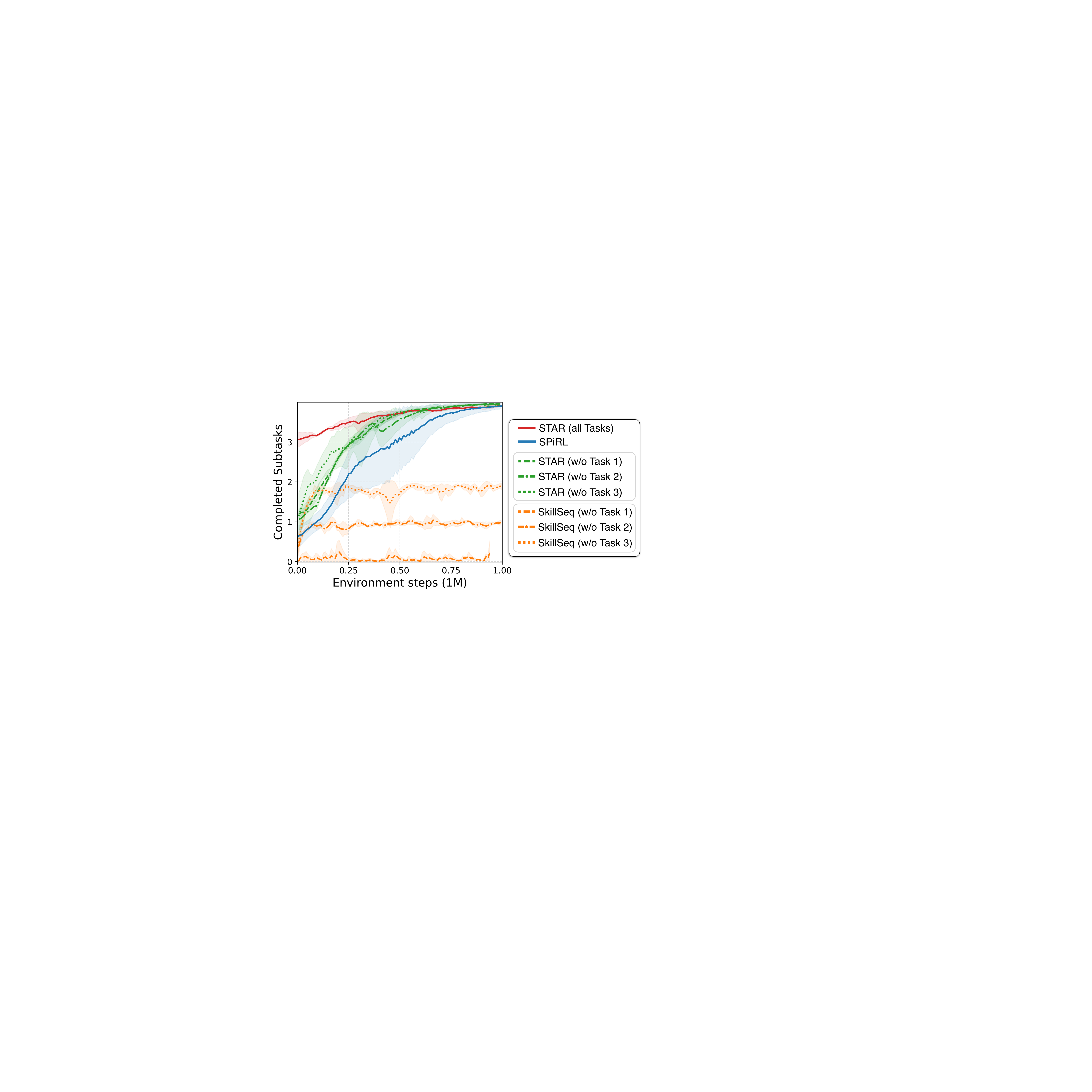}
    \vspace{-0.5cm}
    \caption{Semantic imitation with missing skills in the demonstrations. Our approach \acr~still learns the full task faster than learning without demonstrations (SPiRL), while SkillSeq get stuck at the missing skill.
    }
    \label{fig:robustness_exp}
    \vspace{-0.5cm}
\end{wrapfigure}
In realistic scenarios agents often need to cope with noisy demonstration data, \eg with partial demonstrations or faulty labels. Thus, we test \acr's ability to handle such noise. First, we test imitation from \emph{partial} demonstrations with missing subskills. These commonly occur when there are large differences between source and target domain, \eg the demonstration domain might already have a pot on the stove, and starts with ``turn on the stove'', but in the target domain we need to first place the pot on the stove. We test this in the simulated kitchen tasks by dropping individual subskills from the demonstrations (``w/o Task~$i$' in Figure \ref{fig:robustness_exp}). Figure~\ref{fig:robustness_exp} shows that the SkillSeq approach struggles with such noise: it gets stuck whenever the corresponding skill is missing in the demonstration. In contrast, \acr~can leverage demonstrations that are lacking complete subskills and still learn faster than the no-demonstration baseline SPiRL. When a skill is missing, the \acr~agent finds itself off the demonstration support. Then the objective in~\eqref{eq:policy_objective} regularizes the policy towards the task-agnostic skill prior, encouraging the agent to explore until it finds its way (back) to the demonstration support. This allows our method to bridge ``holes'' in the demonstrations.
We also test \acr's robustness to noisy semantic skill labels, in Section~\ref{sec:noise_results}. We find that \acr~is robust to errors in the annotated skill lengths and to uncertain skill detections. Only frequent, high-confidence mis-detections of skills can lead to erroneous matches and decreased performance. 
Both experiments show that \acr's guidance with semantic demonstrations is robust to noise in the training and demonstration data. %

\section{Conclusion and Limitations}
\label{sec:conclusion}

In this work, we presented \acr, an approach for imitation based on semantic skills that can use cross-domain demonstrations for accelerating RL. \acr~is effective on multiple semantic imitation problems, including using real-world human demonstration videos for learning a robotic kitchen manipulation task. Our results present a promising way to use large-scale human video datasets like EPIC Kitchens~\citep{Damen2021RESCALING} for behavior learning in robotics. %
However, our approach assumes a pre-defined set of semantic skills and semantic skill labels on the training data. We demonstrated how such assumptions can be reduced via the use of pre-trained skill prediction models. 
Yet, obtaining such semantic information from cheaper-to-collect natural language descriptions of the training trajectories without a pre-defined skill set is an exciting direction for future work. Additionally, strengthening the robustness to skill mis-labelings, \eg via a more robust state matching mechanism, can further improve performance on noisy, real-world datasets.

\clearpage
\acknowledgments{
This work was supported by Institute of Information \& Communications Technology Planning \& Evaluation (IITP) grants (No.2019-0-00075, Artificial Intelligence Graduate School Program, KAIST; No.2022-0-00077, AI Technology Development for Commonsense Extraction, Reasoning, and Inference from Heterogeneous Data) and National Research Foundation of Korea (NRF) grant (NRF-2021H1D3A2A03103683), funded by the Korean government (MSIT).
}

\bibliography{bibref_definitions_long,bibtex}

\clearpage

\appendix
\appendix

\section{Full Algorithm}
\label{sec:algorithm}

\begin{algorithm*}[t]
\caption{\acr~(Semantic Transfer Accelerated RL) -- High-level Policy Optimization}
\label{alg:sil}
\begin{algorithmic}[1]
\STATE \textbf{Inputs:} $H$-step reward function $\tilde{r}(s_t, k_t, z_t)$, reward weight $\kappa$, discount $\eta$, target divergences $\delta_q, \delta_p, \delta_l$, learning rates $\lambda_{\pi}, \lambda_Q, \lambda_\alpha$, target update rate $\tau$.
\STATE \textbf{Load pre-trained \& freeze:} low-level policy $\pi^l(a_t \vert s_t, k_t, z_t)$, skill priors $p^\text{demo}(k_t \vert s_t), p^\text{TA}(k_t \vert s_t), p^\text{TA}(z_t \vert s_t, k_t)$, discriminator $D(s)$, progress predictor $P(s)$
\STATE \textbf{Initialize:} replay buffer $\mathcal{D}$, high-level policies $\pi^\text{sem}_\sigma(k_t \vert s_t), \pi^\text{lat}_\theta(z_t\vert s_t, k_t)$, critic $Q_\phi(s_t, k_t, z_t)$, target network $Q_{\bar{\phi}}(s_t, k_t, z_t)$

\STATE

\FOR{each iteration}

\FOR{every $H$ environment steps}
\STATE $k_t \sim \pi^\text{sem}(k_t \vert s_t)$ \COMMENT{sample semantic skill from policy}
\STATE $z_t \sim \pi^\text{lat}(z_t \vert s_t, z_t)$ \COMMENT{sample latent skill from policy}
\STATE $s_{t^\prime} \sim p(s_{t+H} \vert s_t, \pi^l(a_t \vert s_t, k_t, z_t))$ \COMMENT{execute skill in environment}
\STATE $\mathcal{D} \leftarrow \mathcal{D} \cup \{s_t, k_t, z_t, R_\Sigma, s_{t^\prime}\}$ \COMMENT{store transition in replay buffer, $R_\Sigma$ = H-step summed reward}
\ENDFOR

\FOR{each gradient step}
\STATE $\tilde{r} = (1 - \kappa) \cdot R_\Sigma + \kappa \cdot \big[\log D(s_t) - \log\big(1 - D(s_t)\big)\big]$ \COMMENT{compute combined reward}
\STATE $\tilde{r} = \mathds{1}(D(s) < 0.5) \cdot \tilde{r} + \mathds{1}(D(s) \geq 0.5) \cdot \tilde{r} \cdot P(s)$ \COMMENT{optionally shape reward with progress predictor}

\STATE $\bar{Q} = \tilde{r} + \eta \big[ Q_{\bar{\phi}}(s_{t^\prime}, \pi^\text{lat}_\theta(z_{t^\prime} \vert s_{t^\prime}, \pi^\text{sem}_\sigma(k_{t^\prime} \vert s_{t^\prime})))$
\STATE \hspace{4.7cm}$- \big[\alpha_q D_\text{KL}\big(\pi^\text{sem}_\sigma(k_{t^\prime} \vert s_{t^\prime}), p^\text{demo}(k_{t^\prime} \vert s_{t^\prime})\big) \cdot D(s_{t^\prime})$
\STATE \hspace{5cm}$+\;\alpha_p D_\text{KL}\big(\pi^\text{sem}_\sigma(k_{t^\prime} \vert s_{t^\prime}), p^\text{TA}(k_{t^\prime} \vert s_{t^\prime})\big) \cdot \big(1 - D(s_{t^\prime})\big)$
\STATE \hspace{5cm}$+\;\alpha_l D_\text{KL}\big(\pi^\text{lat}_\theta(z_{t^\prime} \vert s_{t^\prime}, k_{t^\prime}), p^\text{TA}(z_{t^\prime} \vert s_{t^\prime}, k_{t^\prime})\big)\big) \big]$ \COMMENT{compute Q-target}

\STATE $(\sigma, \theta) \leftarrow (\sigma,\theta) - \lambda_\pi \nabla_{(\sigma,\theta)} \big[Q_\phi(s_t, \pi^\text{lat}_\theta(z_t \vert s_t, \pi^\text{sem}_\sigma(k_t \vert s_t)))$
\STATE \hspace{4.7cm}$- \big[\alpha_q D_\text{KL}\big(\pi^\text{sem}_\sigma(k_t \vert s_t), p^\text{demo}(k_t \vert s_t)\big) \cdot D(s_t)$
\STATE \hspace{5cm}$+\;\alpha_p D_\text{KL}\big(\pi^\text{sem}_\sigma(k_t \vert s_t), p^\text{TA}(k_t \vert s_t)\big) \cdot \big(1 - D(s_t)\big)$
\STATE \hspace{5cm}$+\;\alpha_l D_\text{KL}\big(\pi^\text{lat}_\sigma(z_t \vert s_t, k_t), p^\text{TA}(z_t \vert s_t, k_t)\big)\big) \big]$ \COMMENT{update high-level policy weights}

\STATE $\phi \leftarrow \phi - \lambda_Q \nabla_\phi \big[ \frac{1}{2}\big(Q_\phi(s_t, k_t, z_t) - \bar{Q} \big)^2 \big]$ \COMMENT{update critic weights}
\STATE $\alpha_q \leftarrow \alpha_q - \lambda_\alpha \nabla_{\alpha_q} \big[ \alpha_q \cdot (D_\text{KL}(\pi^\text{sem}_\sigma(k_t \vert s_t), p^\text{demo}(k_t \vert s_t)) - \delta_q) \big]$ \COMMENT{update alpha values}
\STATE $\alpha_p \leftarrow \alpha_p - \lambda_\alpha \nabla_{\alpha_p} \big[ \alpha_p \cdot (D_\text{KL}(\pi^\text{sem}_\sigma(k_t \vert s_t), p^\text{TA}(k_t \vert s_t)) - \delta_p) \big]$
\STATE $\alpha_l \leftarrow \alpha_l - \lambda_\alpha \nabla_{\alpha_l} \big[ \alpha_l \cdot (D_\text{KL}(\pi^\text{lat}_\theta(z_t \vert s_t, k_t), p^\text{TA}(z_t \vert s_t, k_t)) - \delta_l) \big]$
\STATE $\bar{\phi} \leftarrow \tau \phi + (1 - \tau) \bar{\phi}$ \COMMENT{update target network weights}
\ENDFOR

\ENDFOR
\STATE \textbf{return} trained high-level policies $\pi^\text{sem}_\sigma(k_t \vert s_t), \pi^\text{lat}_\theta(z_t \vert s_t, k_t)$
\end{algorithmic}
\end{algorithm*}

We present a detailed description of the downstream RL algorithm for our \acr~approach in Algorithm~\ref{alg:sil}. It builds on soft actor-critic~\citep{haarnoja2018sac,haarnoja2018sac_algo_applications}. In contrast to the original SAC we operate in a hybrid action space with mixed discrete and continuous actions: $\pi^\text{sem}(k \vert s)$ outputs discrete semantic skill IDs and $\pi^\text{lat}(z \vert s, k)$ outputs continuous latent variables. 

For all input hyperparameters we use the default values from \citet{pertsch2020spirl,pertsch2021skild} and only adapt the regularization weights $\alpha_q, \alpha_p$ and $\alpha_l$ for each task. They can either be set to a fixed value or automatically tuned via dual gradient descent in lines 24-26 by setting target parameters $\delta_q, \delta_p$ and $\delta_l$~\citep{haarnoja2018sac_algo_applications}.

\section{Overview of \citet{pertsch2021skild}}
\label{sec:skild_summary}

While the goal of our work is to imitate semantic skills \emph{across} domains, we build on ideas from \citet{pertsch2021skild}, which use \emph{in-domain} demonstrations. %
\citet{pertsch2021skild} study demonstration-guided RL using a two-layer hierarchical policy architecture: a high-level policy $\pi^{h}(z \vert s)$ outputs temporally extended actions, or \emph{skills}, as learned latent representation $z$. %
The skill $z$ gets decoded into actions $a$ by a learned low-level policy $\pi^{l}(a \vert s, z)$. Here, $z$ captures a skill's behavior in terms of its low-level actions instead of its semantics. 
\citet{pertsch2021skild} assume access to two datasets: demonstration trajectories $\mathcal{D^\text{demo}}$ which solve the task at hand and a task-agnostic dataset $\mathcal{D}^\text{TA}$ of state-action trajectories from a range of prior tasks. %
First, they pre-train the latent skill representation $z$ and the low-level policy $\pi^{l}(a \vert s, z)$ using $\mathcal{D}^\text{TA}$. %
Next, they use the demonstration dataset $\mathcal{D}^{\text{demo}}$ and the task-agnostic dataset $\mathcal{D}^\text{TA}$ to learn a demonstration prior $p^{\text{demo}}(z \vert s)$ and a task-agnostic prior $p^{\text{TA}}(z \vert s)$ over $z$. %
The former captures the distribution over skills in the demonstrations, while the latter represents the skills in the task-agnostic dataset. Additionally, they use both datasets to train a discriminator $D(s)$ to distinguish states sampled from the task-agnostic and demonstration data. 
Both pre-trained prior distributions are used to regularize the high-level policy $\pi^{h}(z \vert s)$ during RL: when $D(s)$ classifies a state as part of the demonstrations, the policy is regularized towards $p^{\text{demo}}(z \vert s)$, encouraging it to imitate the demonstrated skills. In states which $D(s)$ classifies as outside the demonstration support, the policy is regularized towards  $p^{\text{TA}}(z \vert s)$, encouraging it to explore the environment to reach back onto the demonstration support. 
The optimization objective for $\pi^{h}(z \vert s)$ is:
\vspace{-0.2cm}
\begin{equation}
    \mathbb{E}_{\pi^h} \bigg[R(s, a) 
    \underbrace{- \alpha_q D_\text{KL}\big(\pi^h(z \vert s), p^{\text{demo}}(z \vert s)\big) \cdot D(s)}_{\text{demonstration prior regularization}}
    \underbrace{- \alpha_p D_\text{KL}\big(\pi^h(z \vert s), p^{\text{TA}}(z \vert s)\big) \cdot (1-D(s))}_{\text{task-agnostic prior regularization}}\bigg].
\label{eq:skild_objective}
\end{equation}
Crucially, the learned skills $z$ do not represent semantic skills, but instead reflect the underlying sequences of low-level actions. Thus, \citet{pertsch2021skild}'s approach is unsuitable for cross-domain imitation, since a policy would imitate the demonstration's low-level actions instead of its semantics.

\section{Implementation Details}
\label{sec:impl_details}

\subsection{Skill Learning}
\label{sec:impl_details_skill_learning}

We summarize the pre-training objectives of all model components in Table~\ref{tab:models_and_objectives}. We instantiate all components with deep neural networks. We use a single-layer LSTM with 128 hidden units for the inference network and 3-layer MLPs with 128 hidden units for the low-level policy. The skill-representation $z$ is a 10-dimensional continuous latent variable. All skill priors are implemented as 5-layer MLPs with 128 hidden units. The semantic skill priors output logits of a categorical distribution over $k$, the non-semantic prior outputs mean and log-variance of a diagonal Gaussian distribution over $z$. We use batch normalization after every layer and leaky ReLU activations. We auto-tune the regularization weight $\beta$ for training the low-level skill policy using dual gradient descent and set the target to $\SI{2e-2}{}$ for the maze and $\SI{1e-2}{}$ for all kitchen experiments.

When training on image-based human data we add a 6-layer CNN-encoder to the semantic skill prior $p^\text{TA}(k \vert s)$ trained on the source domain dataset $\mathcal{D}_S$. The encoder reduces image resolution by half and doubles the number of channels in each layer, starting with a resolution of 64x64 and 8 channels in the first layer. We use batch normalization and leaky ReLU activations for this encoder too.

The demonstration discriminator $D(s)$ is implemented as a 2-layer MLP with 32 hidden units and no batch normalization to avoid overfitting. We use a sigmoid activation in it's final layer to constrain its output in range $(0, 1)$.

For cross-domain state matching we use a symmetric temporal window with $\gamma^-, \gamma^+ = 0.99$. Only in the experiments with missing skills (see Section~\ref{sec:robustness_exp}) we set $\gamma^-, \gamma^+ = 0$.

All networks are optimized using the RAdam optimizer~\citep{liu2019radam} with parameters $\beta_1 = 0.9$ and $\beta_2= 0.999$, batch size \SI{128}{} and learning rate $\SI{1e-3}{}$. The computational complexity of our approach is comparable to that of prior skill-based RL approaches like~\citet{pertsch2020spirl}. On a single NVIDIA V100 GPU we can train the low-level policy and all skill priors in approximately 10 hours and the demonstration discriminator in approximately 3 hours.

\begin{table*}[t]
\caption{List of all pre-trained model components, their respective functionality and their pre-training objectives. We use $\lfloor \cdot \rfloor$ to indicate stopped gradients and $\tau^T$ to denote demonstration trajectories relabeled with matched target domain states from Section~\ref{sec:matching}.}
\label{tab:models_and_objectives}
\vskip 0.15in
\begin{center}
\begin{small}

\begin{minipage}{1.22\linewidth}
\hspace{-1.8cm}
\begin{tabularx}{\textwidth}{p{2cm}p{1.5cm}p{5cm}X}
\toprule
\textsc{Model} & \textsc{Symbol} & \textsc{Description} & \textsc{Training Objective} \\
\midrule
Skill Policy    & $\pi^l(a \vert s, k, z)$ & Executes a given skill, defined by semantic skill ID and low-level execution latent. & Equation~(1) \\

Demonstration Semantic Skill Distribution    & $p^\text{demo}(k \vert s)$ & Captures semantic skill distribution of demonstration sequences. & $ \mathbb{E}_{s,a,k\sim \tau^T}\bigg[ - \sum_{i=1}^K k_i \cdot \log p^\text{demo}(k_i \vert s) \bigg]$ \\

Task-Agnostic Semantic Skill Prior & $p^\text{TA}(k \vert s)$ & Captures semantic skill distribution of task-agnostic prior experience. & $ \mathbb{E}_{s,a,k\sim \mathcal{D}_T}\bigg[ - \sum_{i=1}^K k_i \cdot \log p^\text{TA}(k_i \vert s) \bigg]$ \\

Task-Agnostic Low-level Execution Prior    & $p^\text{TA}(z \vert s, k)$ & Captures distribution over low-level execution latents from task-agnostic prior experience. &    $ \mathbb{E}_{s,a,k\sim \mathcal{D}_T}\bigg[ D_\text{KL}\big( \lfloor q(z \vert s, a, k) \rfloor, p^\text{TA}(z \vert s, k) \big) \bigg]$     \\

Demonstration Support Discriminator     & $D(s)$ & Determines whether a state is within the support of the demonstrations. & $- \frac{1}{2} \cdot \bigg[\underbrace{\mathbb{E}_{s\sim \tau^T} \big[\log D(s)\big]}_\text{demonstrations} + \underbrace{\mathbb{E}_{s\sim \mathcal{D}_T} \big[\log\big(1 - D(s)\big)\big]}_\text{task-agnostic data}\bigg]$ \\

\bottomrule
\end{tabularx}
\end{minipage}

\end{small}
\end{center}
\vskip -0.1in
\end{table*}

\subsection{Semantic Imitation}
\label{sec:impl_details_imitation}

The high-level policies $\pi^\text{sem}(k \vert s)$ and $\pi^\text{lat}(z \vert s, k)$ are implemented as 5-layer MLPs with batch normalization and ReLU activations. The former outputs the logits of a categorical distribution over $k$, the latter the mean and log-variance of a diagonal Gaussian distribution over $z$. We initialize the semantic high-level policy with the pre-trained demonstration skill prior $p^\text{demo}(k \vert s)$ and the non-semantic high-level policy with the pre-trained task-agnostic latent skill prior $p^\text{TA}(z \vert s, k)$. We implement the critic as a 5-layer MLP with 256 hidden units per layer that outputs a $\vert \mathcal{K} \vert$-dimensional vector of Q-values. The scalar Q-value is then computed as the expectation under the output distribution of $\pi^\text{sem}$.

We use batch size 256, replay buffer capacity of $\SI{1e6}{}$ and discount factor $\gamma = 0.99$. We warm-start training by initializing the replay buffer with \SI{2000}{} steps. We use the Adam optimizer~\cite{kingma2014adam}
with $\beta_1=0.9$, $\beta_2=0.999$ and learning rate $\SI{3e-4}{}$ for updating policy and critic. Analogous to SAC, we train two separate critic networks and compute the $Q$-value as the minimum over both estimates to stabilize training. The target networks get updated at a rate of $\tau = \SI{5e-3}{}$. The latent high-level policy's actions are limited in the range $[-2, 2]$ by a $\tanh$ "squashing function" (see~\citet{haarnoja2018sac}, appendix C). We set all $\alpha$ parameters to fixed values of 10 in the maze navigation task and $\SI{5e-2}{}$ in all kitchen tasks.

For reward computation we set the factor $\kappa = 0.9$, \ie we blend environment and discriminator-based rewards. In practice, we find that we can improve convergence speed by using a \emph{shaped} discriminator reward that increases towards the end of the demonstration. This is comparable to goal-proximity based rewards used in in-domain imitation, \eg~\citet{lee2021generalizable}. To compute the shaped reward, we pre-train a progress predictor $P(s)$ along with the discriminator $D(s)$. $P(s)$ estimates the time step of a state within a demonstration relative to the total length of the demonstration, thus its outputs are bound in the range $[0, 1]$. We implement the progress predictor as a simple 3-layer MLP with a sigmoid output activation. During RL training we can then compute the shaped reward as:
\begin{align}
r(s, a) &= \kappa \cdot R(s, a) + (1 - \kappa) \cdot 
\begin{cases}
P(s) \cdot R_D & \text{if } P(s) \geq 0.5 \\
R_D & \text{otherwise}
\end{cases}\nonumber \\
&\text{with } R_D = \log D(s_t) - \log\big(1 - D(s_t)\big)
\end{align}

For all RL results we average the results of three independently seeded runs and display mean and standard deviation across seeds. The computation time for these experiments varies by environment and is mainly determined by the simulation time of the used environments. Across all environments we can typically finish downstream task training within <12h on a single NVIDIA V100 GPU.

\subsection{Comparisons}
\label{sec:impl_details_comparisons}

\paragraph{SPiRL.} We follow the approach of~\citet{pertsch2020spirl} which first trains a latent skill representation from task-agnostic data and then uses a pre-trained task-agnostic prior to regularize the policy during downstream learning. To allow for fair comparison, we adapt the SPiRL approach to work with our semantic skill model. In this way both SPiRL and \acr~use the same set of learned semantic skills. During downstream task learning we regularize both high-level policies $\pi^\text{sem}(k \vert s)$ and $\pi^\text{lat}(z \vert s, k)$ using the corresponding task-agnostic skill priors $p^\text{TA}(k \vert s)$ and $p^\text{TA}(z \vert s, k)$, analogous to the task-agnostic skill prior regularization in the original SPiRL work.

\paragraph{SkiLD.} We similarly adapt SkiLD~\citep{pertsch2021skild} to work with our learned semantic skill model. In contrast to the SPiRL comparison, we now regularize both high-level policies with skill distributions trained on the target domain demonstrations whenever $D(s)$ classifies a state as being part of the demonstration support (see Section~\ref{sec:skild_summary}).

\paragraph{SkillSeq.} We pre-train a skill-ID conditioned policy on the task-agnostic target domain dataset $\mathcal{D}_T$ using behavioral cloning. We split this policy into a 3-layer MLP encoder and a 3-layer MLP policy head that produces the output action. The policy has an additional 3-layer MLP output head that is trained to estimate whether the current skill terminates in the input state. We use the semantic skill labels $k$ in the task-agnostic dataset to determine states in which a skill ends and train the termination predictor as a binary classifier. During downstream learning, we use a programmatic high-level policy that has access to the true sequence of semantic skills required to solve the downstream task and conditions the low-level policy on these skill IDs one-by-one. The skill ID is switched to the next skill when the pre-trained termination predictor infers the current state as a terminal state for the current skill. For fair comparison we use online RL for finetuning the skill-conditioned policy via soft actor-critic (SAC, ~\citet{haarnoja2018sac}).

\paragraph{BC+RL.} We train a policy directly on the source domain demonstrations via behavioral cloning. We then use this pre-trained policy to initialize the policy during target task training in the target domain. We fine-tune this initialization using SAC with the rewards provided by the target environment. Similar to~\citet{rajeswaran2018learning,nair2018overcoming} we regularize the policy towards the pre-trained BC policy during downstream learning.

\begin{figure*}[t]
    \centering
    \includegraphics[width=1.0\linewidth]{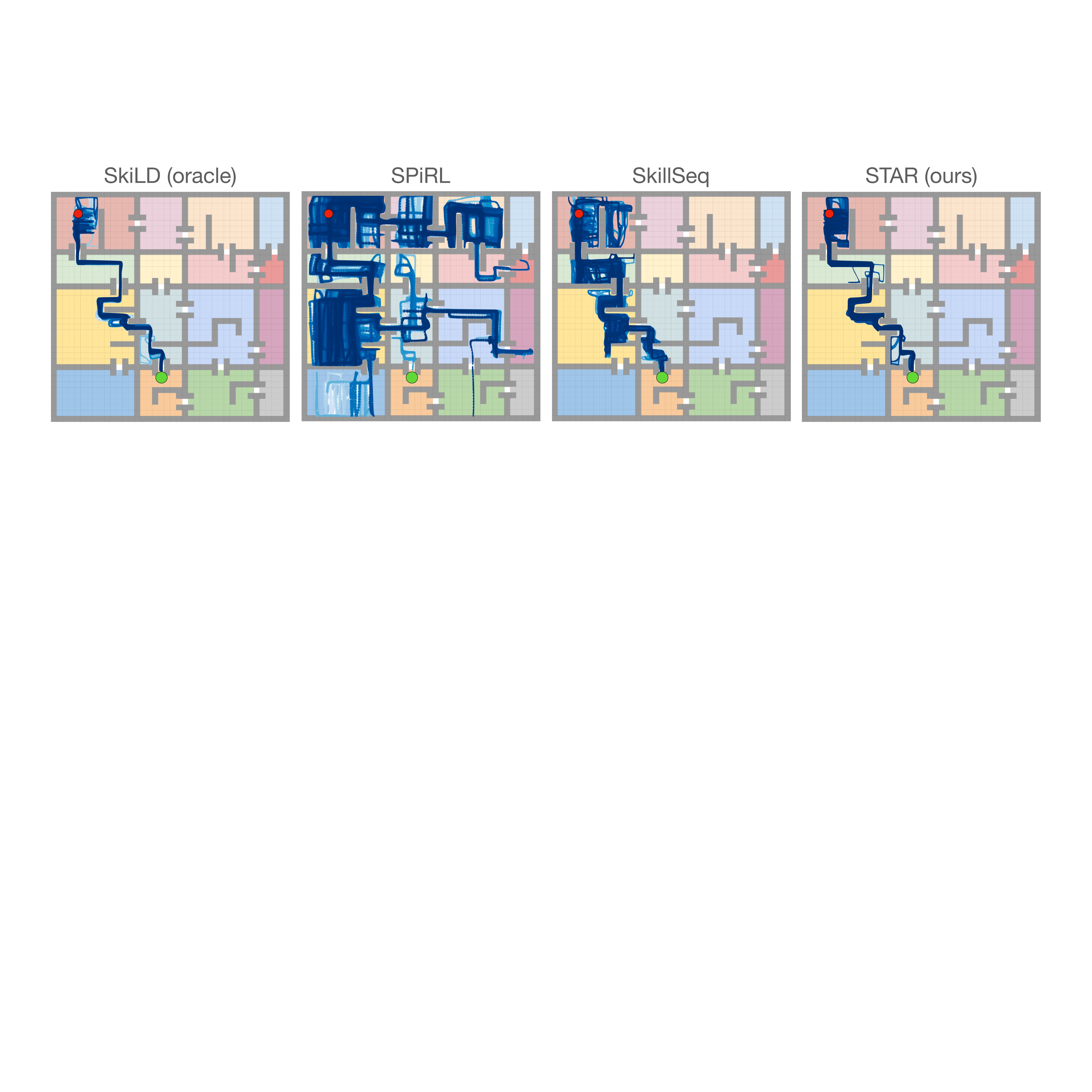}
    \caption{Qualitative maze results. We visualize the trajectories of the different policies during training. The \textbf{SkiLD} approach leverages in-domain demonstrations to quickly learn how to reach the goal. \textbf{SPiRL} leverages skills from the task-agnostic dataset to widely explore the maze, but fails to reach the goal. \textbf{SkillSeq} makes progress towards the goal, but can get stuck in intermediate rooms, leading to a substantially lower success rate than the oracle method. Our approach, \textbf{\acr}, is able to match the performance of the oracle baseline and quickly learn to reach the goal while following the sequence of demonstrated rooms. 
    }
    \label{fig:quali_maze_results}
\end{figure*}

\subsection{Environments and Data Collection}
\label{sec:details_env_and_data}

\begin{wrapfigure}{r}{0.4\textwidth}
    \centering
    \includegraphics[width=1.0\linewidth]{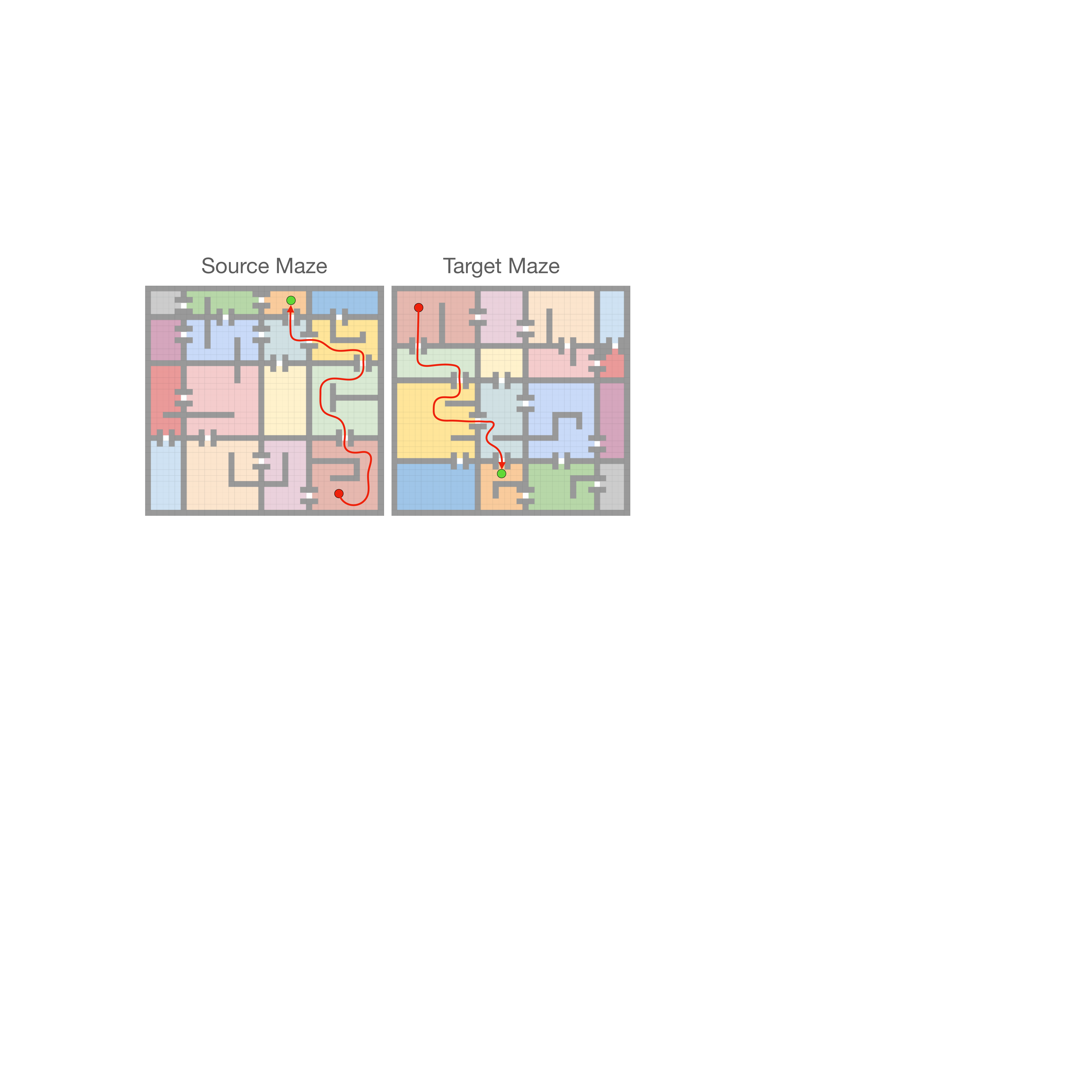}
    \caption{Source and target semantic maze navigation environments. A room's color indicates its semantic ID. The red trajectory shows the traversal of semantic rooms demonstrated in the source domain and the corresponding trajectory in the target domain. The low-level planar velocity commands required to follow the demonstration in the target domain is substantially different.
    }
    \label{fig:maze_envs}
\end{wrapfigure}

\paragraph{Maze navigation.} We generate two maze layouts with the same number of rooms. We indicate a room's semantic ID via its color in Figure~\ref{fig:maze_envs}. We ensure the same ``room connectivity'' between both layouts, \ie corresponding semantic rooms have the same openings between each other. For example the yellow room connects to the blue room but not to the green room in both layouts. This ensures that we can follow the same sequence of semantic rooms in both environments. While we ensure that the semantic layout of the mazes is equivalent, their physical layout is substantially different: the mazes are rotated by 180 degrees, for example the red room is in the bottom right corner for the first maze but in the top left corner for the second. Additionally, the layout of individual rooms and the positions of obstacle walls change between the domains. As a result, simple imitation of the low-level planar velocity actions from one domain will not lead to successfully following the same sequence of semantic rooms in the other domain. We define a total of 48 semantic skills: one for each room-to-room traversal, \eg ``go from the red room to the green room'', and one for reaching a goal within each room, \eg ``reach a goal in the green room''. Thus, the semantic description of a demonstrated trajectory could for example be: ``Go from the red room to the green room, then from the green to the beige room, $\dots$, then from the blue to the orange room and then reach a goal in the orange room.''

\begin{figure}[t]
    \centering
    \includegraphics[width=1.0\linewidth]{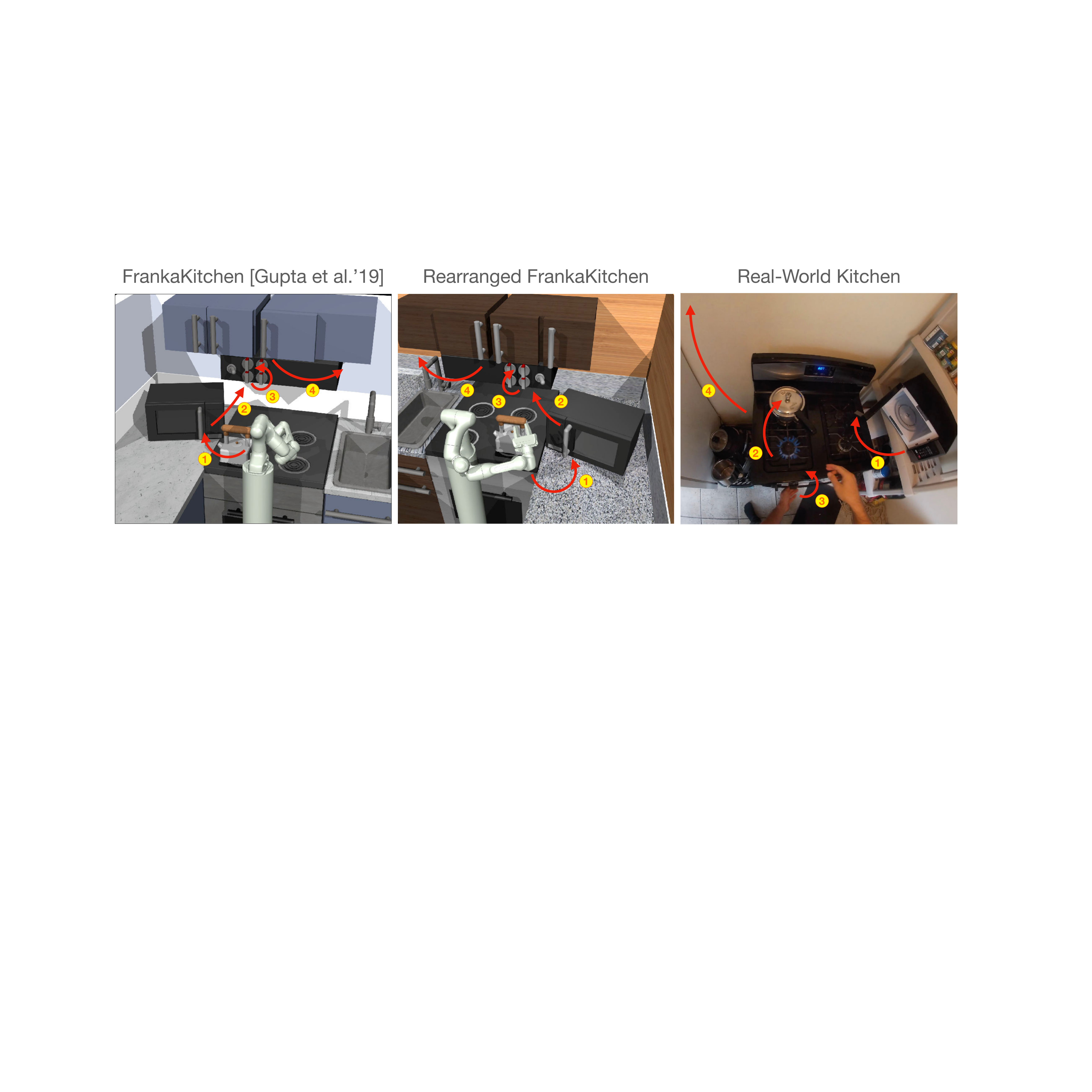}
    \caption{Three semantically equivalent kitchen environments. \textbf{Left}: FrankaKitchen environment~\citep{gupta2019relay}, \textbf{middle}: rearranged FrankaKitchen environment, \textbf{right}: real-world kitchen. In all three environments we define the same set of seven semantic object manipulation skills like ``open the microwave'', ``turn on the stove'' etc. The two simulated kitchen environments require different robot joint actuations to perform the same semantic skills. The real-world kitchen has a different agent embodiment (robot vs. human), layout and observation domain (low-dimensional state vs image observations).
    }
    \label{fig:kitchen_envs}
\end{figure}

\paragraph{Simulated Kitchen.} We use the FrankaKitchen environment of \citet{gupta2019relay} (see Figure~\ref{fig:kitchen_envs}, left) and define a set of seven semantic manipulation skills: opening the microwave, opening the slide and hinge cabinet, turning on bottom and top stove and flipping the light switch. We also create a rearranged version of the kitchen environment (Figure~\ref{fig:kitchen_envs}, middle) with different layout and visual appearance but the same set of semantic interaction options. In both environments we use the state-action definition of \citet{gupta2019relay}: (1)~a 60-dimensional state representation consisting of the agent's joint state as well as object states like opening angles or object pose, (2)~a 9-dimensional action space consisting of 7 robot joint velocities and two gripper finger positions.

For the FrankaKitchen environment we can use the data provided by \citet{gupta2019relay}: 600 human teleoperated sequences each solving four different semantic tasks in sequence. In the newly created rearranged kitchen environment we collect a comparable dataset by controlling the robot via trajectory optimization on a dense reward function. We use the CEM implementation of \citet{Lowrey-ICLR-19}. For both datasets we label the semantic skills by programmatically detecting the end state of an object interaction using the low-dimensional state representation.

\paragraph{Real-World Kitchen.} 
Data collection is performed by fixating a GoPro camera to the head of a human data collector which then performs a sequence of semantic skills. The camera is angled to widely capture the area in front of the human. During data collection and within each trajectory we vary the viewpoint, skill execution speed and hand used for skill execution. 
We collect 20 human demonstration sequences for the task sequence: open microwave, move kettle, turn on stove, open cabinet. We then automatically generate semantic skill predictions via zero-shot inference with a pre-trained action recognition model. Specifically, we use the publicly available SlowFast model trained on the EPIC Kitchens 100 dataset~\citep{Damen2021RESCALING,fan2020pyslowfast}. The model takes in a window of 32 consecutive video images at a $256\times256$ px resolution and outputs a distribution over 97 verb and 300 object classes. Since our simulated FrankaKitchen target environment does not support the same set of skills, we define a mapping from the output of the EPIC Kitchens model to the applicable skills in the Franka Kitchen environment, \eg we will map outputs for the verb ``open'' and the noun ``microwave'' to the ``open microwave'' skill in FrankaKitchen. Note that some skill distinctions in the FrankaKitchen environment are not supported by EPIC Kitchens, like ``turn on top burner'' vs ``turn on bottom burner''. In such cases we map the outputs of the EPIC Kitchens model to a single skill in the target environment. With this skill mapping we finetune the EPIC Kitchens model for outputting the relevant classes. Note that this model finetuning is performed with the original EPIC Kitchens data, \ie \emph{no} additional, domain specific data is used in this step and no additional annotations need to be collected. This finetuning is performed such that the resulting model directly outputs a distribution over the relevant skills. Alternatively, the relevant skills could be extracted from the output of the original model and the distribution could be renormalized.

To generate the skill predictions for the human video demonstrations, we move a sliding window of 32 frames over the demonstrations and generate a prediction in each step using the EPIC Kitchens model. We pad the resulting skill distribution sequence with the first and last predicted skill distribution to obtain the same number of skill predictions as there are frames in the demonstration video. Then we use the sequence of skill distributions to perform cross-domain matching and semantic imitation as detailed in Section~\ref{sec:star_core}, without any changes to the algorithm.

\section{Imitation Learning Results}
\label{sec:imitation_results}

\begin{wrapfigure}{r}{0.4\textwidth}
    \centering
    \vspace{-1.5cm}
    \includegraphics[width=1.0\linewidth]{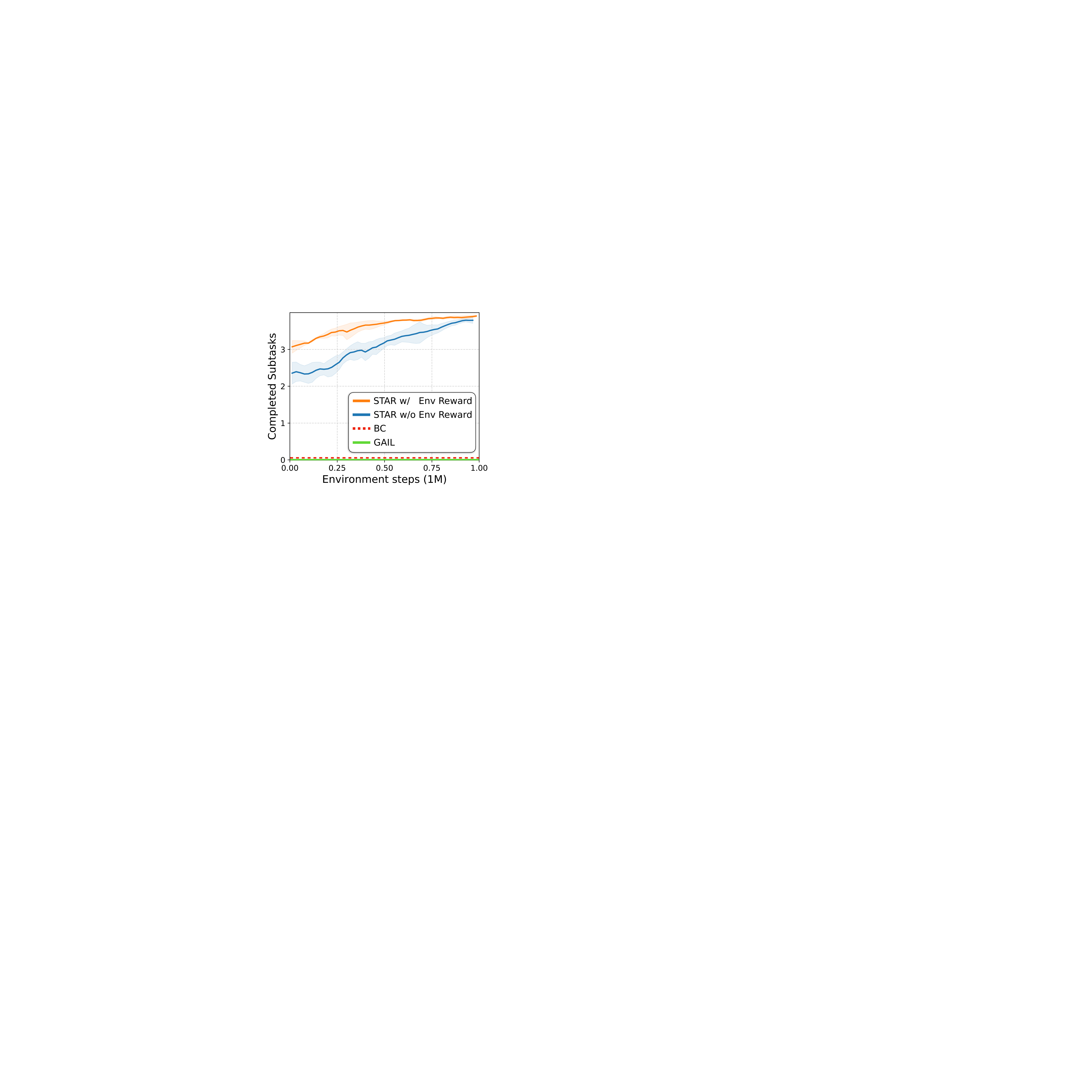}
    \caption{Imitation learning on the simulated FrankaKitchen task. Our approach STAR is able to learn the target task even without access to any environment rewards, while common imitation learning approaches fail to learn the task due to the large domain gap between source demonstrations and target task environment.
    }
    \label{fig:il_results}
    \vspace{-1cm}
\end{wrapfigure}

We evaluate our approach in the ``pure'' imitation learning setting in the kitchen environment. Here, we assume \emph{no access} to environment rewards. Instead, we rely solely on the discriminator-based reward learned from the cross-domain demonstrations to guide learning (see Section~\ref{sec:star_core}). We present evaluations in the FrankaKitchen environment in Figure~\ref{fig:il_results}. Our approach STAR is able to learn the target task from demonstrations without any environment rewards, although learning is somewhat slower than in the demonstration-guided RL setting with environment reward access. In contrast, standard imitation learning approaches are unable to learn the task since they struggle with the large domain gap between source domain demonstrations and target domain execution. These results show that our approach STAR is applicable both, in the demonstration-guided RL setting \emph{with} environment rewards, and in the imitation learning setting \emph{without} environment rewards.

\section{Label Noise Robustness Analysis}
\label{sec:noise_results}

An important aspect for the scalability of an approach is its ability to cope with noise in the training data. While prior work on skill-based RL has investigated the robustness of such approaches to suboptimal behavior in the training data~\citep{pertsch2020spirl}, we will focus on an aspect of the training data that is specifically important for our cross-domain imitation approach: the semantic skill labels. In this section, we investigate the robustness of our approach to different forms of noise on the semantic skill labels. Such noise can either be introduced through inaccuracies in the manual labeling process or via an automated form of skill labeling, as performed with the EPIC kitchens models in Section~\ref{sec:human-demo}. To cleanly investigate the robustness to \emph{different forms} of skill label noise, we start from a noise-free set of labels, which we can easily obtain programmatically in the simulated FrankaKitchen environment. We then artificially perturb the labels to introduce artifacts that mimic realistic labeling errors. This allows us to (1)~investigate different forms of noise independently and (2)~vary the magnitude of the introduced noise in a controlled way.

Specifically, we introduce noise along three axis:
\begin{itemize}
    \item \textbf{skill length noise}: artificially perturbs the length of a labeled skill within a range $[1 - l_n \dots 1+l_n]$ of the true length of the skill, mimicking a labeler's uncertainty on when exactly a skill ends
    
    \item \textbf{skill uncertainty noise}: perturbs the distribution over detected skills around $N_n$ transition between skills by adding probability weight to erroneous skills produced via a random walk, mimicking the uncertainty \eg produced by a pre-trained action recognition model
    
    \item \textbf{skill misdetection noise}: adds $N_m$ incorrectly detected skill segments at randomly sampled points throughout the sequence of randomly sampled lengths, mimicking mis-labelings which can (rarely) occur in human data or (more frequently) in auto-labeled data
\end{itemize} 

\begin{figure}[t]
    \centering
    \includegraphics[width=1.0\linewidth]{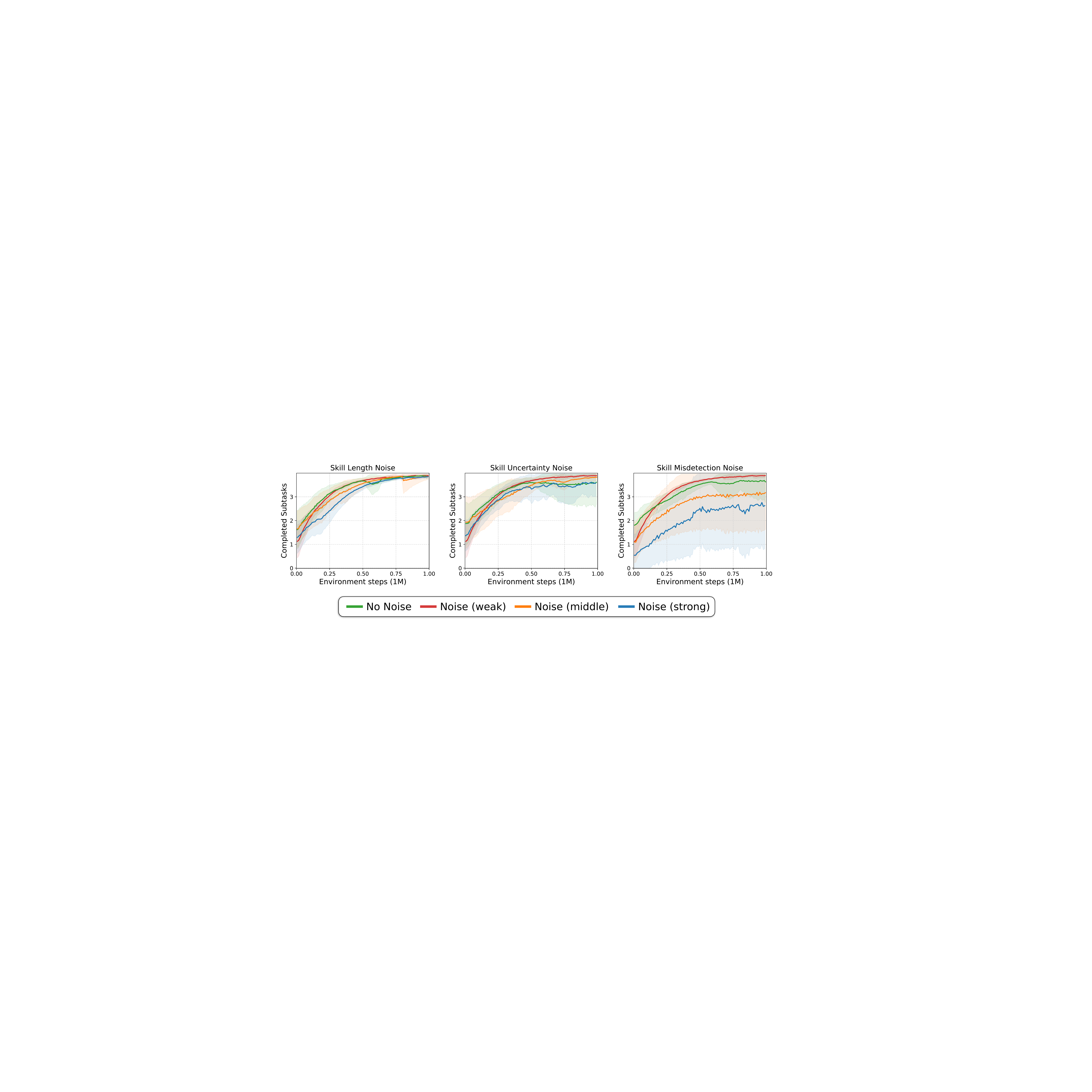}
    \caption{Robustness of our approach, STAR, to different forms of noise in the semantic skill labels. Our approach is robust to noise in the length of annotated skills and uncertainty between different skills. While STAR is also shows some robustness to completely incorrect skill labels, frequent and confident mis-detections / mis-labelings can lead to errors during the cross-domain matching and thus impair learning performance on the target task.
    }
    \label{fig:noise_results}
\end{figure}

\begin{table}[t]
\caption{Parametrization of the noise levels for the skill label robustness experiment.}
\label{tab:noise_levels}
\vskip 0.15in
\begin{center}
\begin{small}

\begin{tabularx}{\textwidth}{lccc}
\toprule
& \textsc{Skill Length Noise} & \textsc{Skill Uncertainty Noise} & \textsc{Skill Misdetection Noise} \\
\midrule
\makecell{Varied\\Parameter}   & \makecell{$l_n$\\(Percentual length\\noise window)} & \makecell{$N_n$\\(Number of\\uncertain segments)} & \makecell{$N_m$\\(number of\\misdetected segments)}\\

\midrule

Weak Noise    & 10 \% & 1 & 1 \\

Middle Noise    & 20 \% & 2 & 2 \\

Strong Noise    & 30 \% & 3 & 3 \\

\bottomrule
\end{tabularx}

\end{small}
\end{center}
\vskip -0.1in
\end{table}

We show evaluations of our approach with different \emph{levels} of noise along all three axis in Figure~\ref{fig:noise_results}. We perform these evaluations in the simulated FrankaKitchen environment and average performance across 10 seeds to reduce the noise-induced variance in the results. The parameters of the different tested noise levels are detailed in Table~\ref{tab:noise_levels}.

The results in Figure~\ref{fig:noise_results} show that STAR is robust to a wider range of noise levels in the annotated skill length and uncertainty between the skills: the performance does not significantly change even with increased noise levels. However, we find that confident mis-predictions / mis-labelings of skills can have a negative impact on the performance. Particularly if mis-predictions happen frequently (``Noise (strong)''), states between the source and target domain can be mismatched, leading to worse target task performance. But we find that even in the case of mis-detections STAR is able to handle a moderate amount of such noise robustly, which is important for STAR's scalability to large and noisy real-world datasets.

\newpage

\section{Detailed Ablation Description}
\begin{figure}[t]
    \centering
    \includegraphics[width=0.7\linewidth]{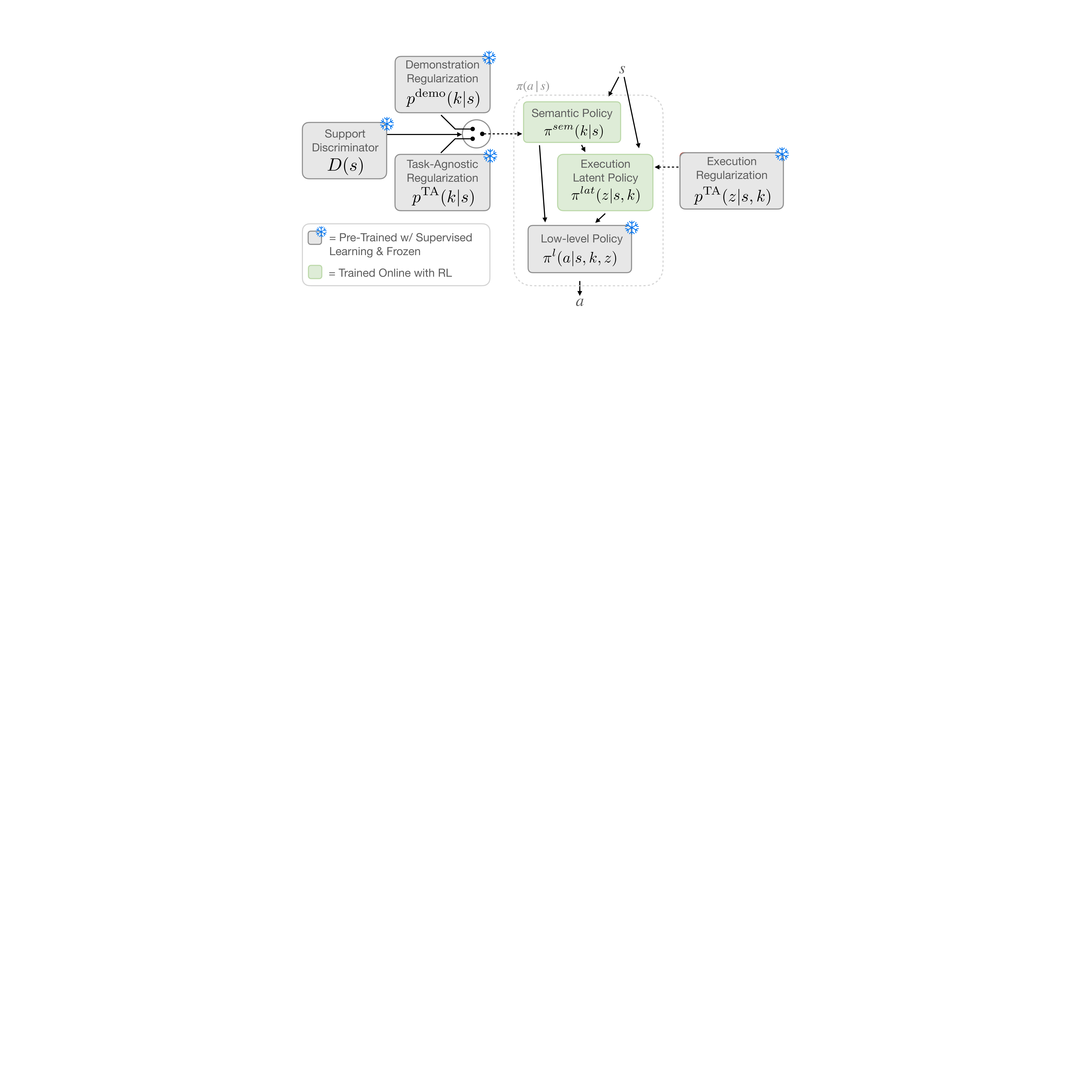}
    \caption{Visualization of all model components. The colors indicate the objective type used for training. Only the high-level policy is trained with online RL on the downstream task, all other components are pre-trained fully offline via supervised learning and frozen during downstream training.}
    \label{fig:component_overview}
\end{figure}
\begin{figure}[t]
    \hspace{-2.5cm}
    \includegraphics[width=1.32\linewidth]{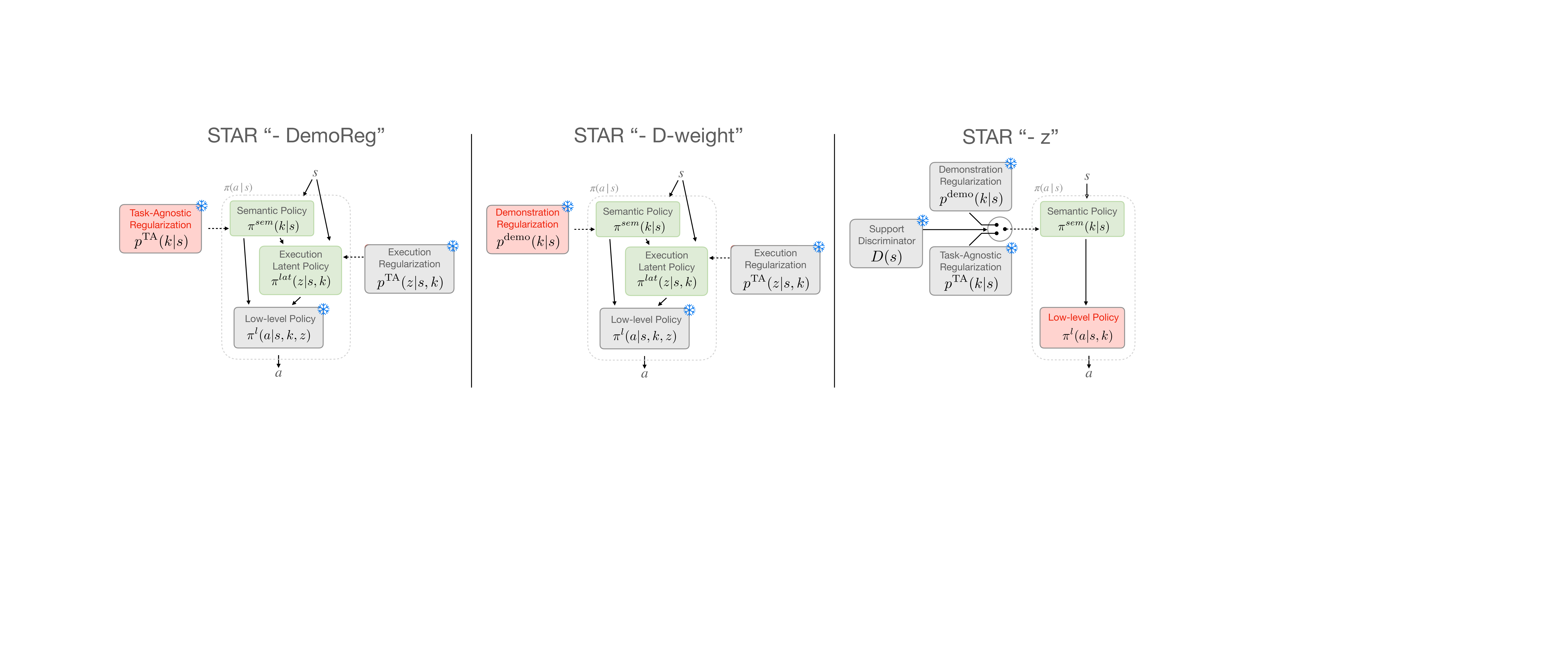}
    \caption{Overview of some of the performed ablations. \textbf{-DemoReg}: removes the demonstration regularization for the high-level semantic policy and uses only the task-agnostic prior for regularization, \textbf{-D-weight}: removes the discriminator-based weighting between demonstration and task-agnostic regularization and uses only the former for guiding the policy during downstream training, \textbf{-z}: removes the latent variable $z$ from the low-level policy and instead uses a deterministic low-level policy and \emph{no} execution latent policy.}
    \label{fig:ablation_overview}
\end{figure}

We provide an overview of the components of our approach in Figure~\ref{fig:component_overview}. The figure highlights that most components are trained offline with simple supervised objectives and then frozen during downstream task learning, making their training straightforward and reproducible. Only the high-level semantic and execution policy are trained via online RL on the downstream task.

We also provide a more detailed description of the performed ablation studies from Figure~\ref{fig:quant_results}, right, below. These ablation studies demonstrate the importance of the different components of our model. Finally, we visualize the resulting models for multiple of our ablation studies in Figure~\ref{fig:ablation_overview}.

\paragraph{STAR - D-reward.} Ablates the discriminator-based dense reward (see Section~\ref{sec:star_core}). Instead trains the high-level policy only based on the environment-provided reward on the downstream task.
\paragraph{STAR - TempAgg.} Ablates the temporal aggregation during cross-domain matching (see Section~\ref{sec:matching}). Instead uses single state semantic skill distributions to find matching states.
\paragraph{STAR - DemoReg.} Ablates the policy regularization with cross-domain skill distributions. Instead simply regularizes with task-agnostic skill priors derived from the target domain play data (see Figure~\ref{fig:ablation_overview}, left).
\paragraph{STAR - D-weight.} Ablates the discriminator-based weighting between demonstration and task-agnostic skill distributions. Instead always regularizes the high-level semantic policy towards the demonstration skill distribution (see Figure~\ref{fig:ablation_overview}, middle).
\paragraph{STAR - z.} Ablates the use of the latent execution variable $z$ in the skill policy. Instead trains a simpler low-level policy \emph{without} latent variable $z$ and removes the execution latent policy (see Figure~\ref{fig:ablation_overview}, right).

\section{Additional Ablation Experiments}
We perform an additional ablation experiment to test whether replacing the high-level policy's weighted KL-regularization scheme from equation~\ref{eq:policy_objective} with a simpler behavioral cloning regularization objective can lead to comparable performance. Concretely, we replace the policy's objective from equation~\ref{eq:policy_objective} with:
\begin{align}
    \max_{\pi^h} \bigg[Q(s, a) 
    \underbrace{- \alpha \mathbb{E}_{k \sim \pi^\text{sem}(k \vert s)} p^{\text{demo}}(k \vert s)}_{\text{BC regularization}}
    \underbrace{- \alpha_l D_\text{KL}\big(\pi^\text{lat}(z \vert s, k), p^\text{TA}(z \vert s, k)\big)}_{\text{task-agnostic execution prior regularization}}\bigg].
    \label{eq:bc_reg_objective}
\end{align}
We also experimented with removing the execution prior regularization term, \ie setting $\alpha_l = 0$, but found it to be crucial for training since the initial policy rapidly degrades without it.

\begin{wrapfigure}{r}{0.4\textwidth}
    \centering
    \label{fig:bc_reg_exp}
    \includegraphics[width=\linewidth]{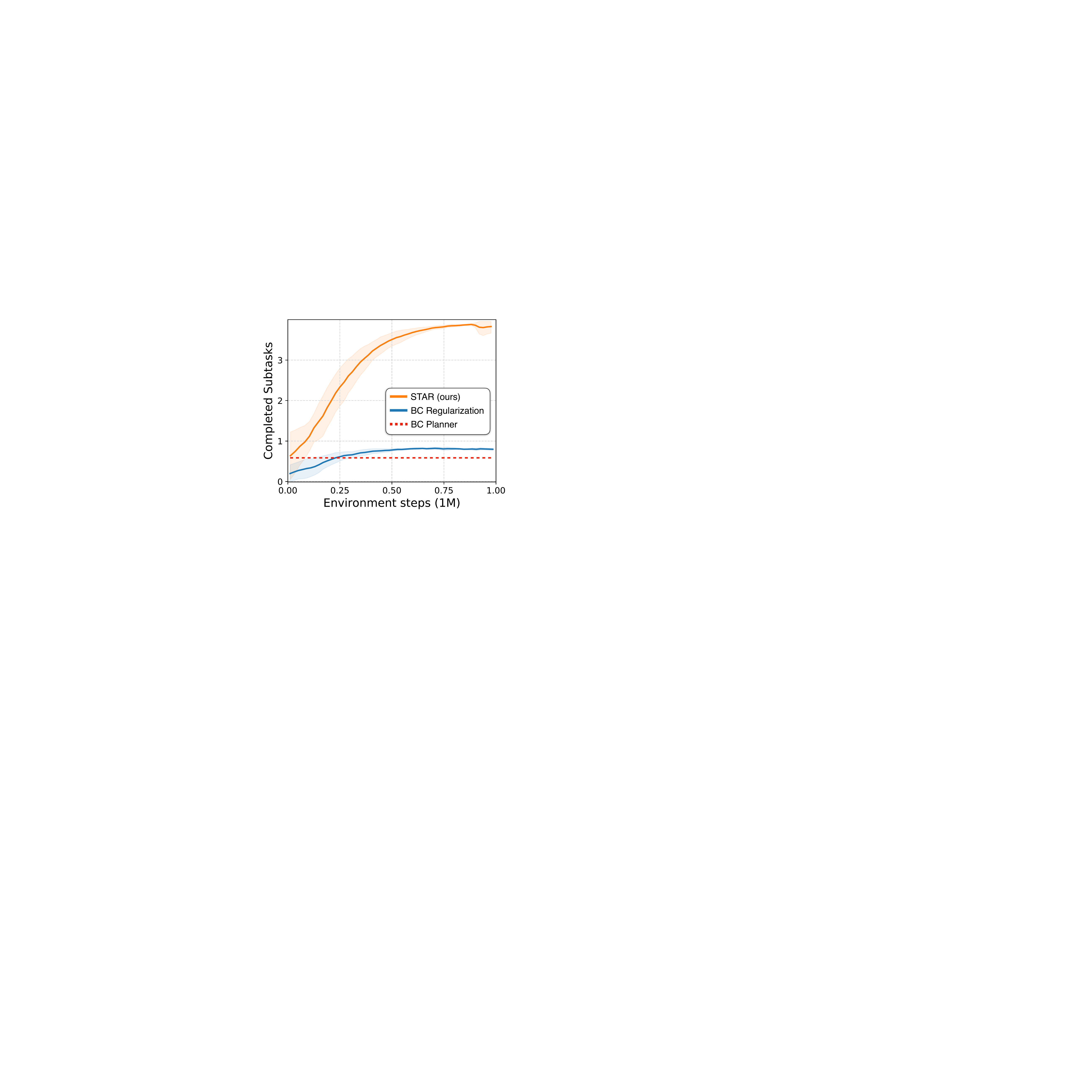}
    \vspace{-1cm}
\end{wrapfigure}
We report quantitative results on the human video demonstration to simulated kitchen manipulation task in the figure on the right. The \emph{BC-Reg} objective in equation~\ref{eq:bc_reg_objective} obtains 75\% lower performance than our full objective from equation~\ref{eq:policy_objective}. This is because the behavioral cloning regularization is also computed on states outside the demonstrations' support, leading to incorrect regularization. Instead, our approach uses the discriminator to only apply regularization \emph{within} the support of the demonstrations.

We also add comparison to an even simpler baseline that clones the transferred semantic skill embeddings from the demonstrations, equivalent to a \textbf{semantic-level BC planner}. This approach does not perform well due to accumulating errors of the high-level planner (see figure on the right). Without online training, this approach cannot correct the shortcomings of the planner.

\end{document}